\documentclass[letterpaper]{article} 
\usepackage{aaai24}  
\usepackage{times}  
\usepackage{helvet}  
\usepackage{courier}  
\usepackage[hyphens]{url}  
\usepackage{graphicx} 
\usepackage{natbib}  
\usepackage{caption} 
\usepackage{algorithm}

\usepackage{newfloat}
\usepackage{listings}
\DeclareCaptionStyle{ruled}{labelfont=normalfont,labelsep=colon,strut=off} 
\floatstyle{ruled}
\newfloat{listing}{tb}{lst}{}
\floatname{listing}{Listing}
\title{AAAI Press Anonymous Submission\\Instructions for Authors Using \LaTeX{}}
\author{
    Written by AAAI Press Staff\textsuperscript{\rm 1}\thanks{With help from the AAAI Publications Committee.}\\
    AAAI Style Contributions by Pater Patel Schneider,
    Sunil Issar,\\
    J. Scott Penberthy,
    George Ferguson,
    Hans Guesgen,
    Francisco Cruz\equalcontrib,
    Marc Pujol-Gonzalez\equalcontrib
}

\usepackage{bibentry}

\usepackage[utf8]{inputenc} 
\usepackage[T1]{fontenc}    
\usepackage{url}            
\usepackage{booktabs}       
\usepackage{amsfonts}       
\usepackage{nicefrac}       
\usepackage{microtype}      
\usepackage{xcolor}         
\usepackage{microtype}
\usepackage{graphicx}
\usepackage{subfigure}
\usepackage{booktabs} 
\usepackage{multicol}

\usepackage{times}
\usepackage{soul}
\usepackage{url}
\usepackage{graphicx}
\usepackage{amsmath}
\usepackage{amsthm}
\usepackage{algorithm}
\usepackage[noend]{algorithmic}
\usepackage{amsfonts,amssymb} 

\usepackage[capitalize,noabbrev]{cleveref}

\usepackage{bm,array}
\usepackage{comment}
 \usepackage{makecell}
 \usepackage{color}
 \usepackage{multirow}
 \usepackage{float}

\newcommand{\cA}{{\mathcal{A}}}

\newcommand{\cS}{{\mathcal{S}}}

\newcommand{\cX}{{\mathcal{X}}}
\newcommand{\cK}{{\mathcal{K}}}

\newcommand{\cN}{{\mathcal{N}}}

\newcommand{\cT}{{\mathcal{T}}}

\newcommand{\cH}{{\mathcal{H}}}

\newcommand{\cO}{{\mathcal{O}}}

\newcommand{\cB}{{\mathcal{B}}}

\newcommand{\bPi}{\pmb{\Pi}}

\newcommand{\bbE}{\mathbb{E}}

\newcommand{\KL}{\textsc{KL}}

\newcommand{\A}{\alpha}
\newcommand{\E}{\textit{e}}

\theoremstyle{plain}
\newtheorem{theorem}{Theorem}
\newtheorem{proposition}[theorem]{Proposition}
\newtheorem{lemma}[theorem]{Lemma}
\newtheorem{corollary}[theorem]{Corollary}
\theoremstyle{definition}

\theoremstyle{remark}

\newcommand{\squishlist}{
 \begin{list}{$\bullet$}
  { \setlength{\itemsep}{0pt}
     \setlength{\parsep}{2pt}
     \setlength{\topsep}{2pt}
     \setlength{\partopsep}{0pt}
     \setlength{\leftmargin}{1.5em}
     \setlength{\labelwidth}{1em}
     \setlength{\labelsep}{0.5em} } }

\newcommand{\squishend}{
  \end{list}  }

\newif\ifnotes\notestrue
\def\htien#1{}

\def\cment#1{{\color{blue}{#1}}}

\usepackage{graphicx}
\usepackage{multirow}
\newcommand{\showinstance}[4][0.25]{
\begin{minipage}{#1\textwidth}
    \centering
    \includegraphics[width=1.0\textwidth,trim={#2 0 0 0},clip]{graphs/#3.pdf}
    \caption*{\detokenize{#4}}
    \captionsetup{justification=centering}
\end{minipage}
}
\newcommand{\showlegend}[3][0.25]{
\begin{minipage}{#1\textwidth}
    \centering
    \includegraphics[width=1.0\textwidth,trim={#2cm 0 0 0},clip]{graphs/#3.pdf}
\end{minipage}
}

\usepackage[textsize=tiny]{todonotes}

\title{Mimicking To Dominate: Imitation Learning Strategies for Success in Multiagent Competitive Games}
\author {
	Viet The Bui\textsuperscript{\rm 1},
	Tien Mai\textsuperscript{\rm 1},
	Thanh H. Nguyen\textsuperscript{\rm 2},
}
\affiliations{
	\textsuperscript{\rm 1}Singapore Management University
	Singapore, Singapore\\
		\textsuperscript{\rm 2}University of Oregon
		Eugene, Oregon, United States\\
{	tvbui@smu.edu.sg; atmai@smu.edu.sg; thanhhng@cs.uoregon.edu}
}

\begin{document}

\maketitle

\begin{abstract}
Training agents in multi-agent competitive games presents significant challenges due to their intricate nature. These challenges are exacerbated by dynamics influenced not only by the environment but also by opponents' strategies. Existing methods often struggle with slow convergence and instability. To address this, we harness the potential of imitation learning to comprehend and anticipate opponents' behavior, aiming to mitigate uncertainties with respect to the game dynamics. Our key contributions include: (i) a new multi-agent imitation learning model for predicting next moves of the opponents --- our model works with hidden opponents' actions and local observations; (ii) a new multi-agent reinforcement learning algorithm that combines our imitation learning model and policy training into one single training process; and (iii) extensive experiments in three challenging game environments, including an advanced version of the Star-Craft multi-agent challenge (i.e., SMACv2). Experimental results show that our approach achieves superior performance compared to existing state-of-the-art multi-agent RL algorithms.


    

\end{abstract}
\section{Introduction}
Recent works in multi-agent reinforcement learning (MARL) have made a significant progress in developing new effective MARL algorithms that can perform well in complex multi-agent environments including SMAC (StarCraft Multi-Agent Challenge)~\cite{yu2022surprising,samvelyan2019starcraft}. Among these works, centralized training and decentralized execution (CTDE)~\cite{foerster2018counterfactual} has attracted a significant attention from the RL research community due to its advantage of leveraging global information to train a centralized critic (aka. actor-critic methods~\cite{lowe2017multi}) or a joint Q-function (aka. value-decomposition methods~\cite{rashid2020monotonic,sunehag2017value}). This approach enables a more efficient and stable learning process while still allowing agents to act in a decentralized manner. Under this CTDE framework, off-policy methods such as MADDPG~\cite{lowe2017multi} and QMIX~\cite{rashid2020monotonic} have become very popular as a result of their data efficiency and state-of-the-art (SOTA) results on a wide range of benchmarks. 

On the other hand, on-policy gradient methods have been under-explored so far in MARL due to their data consuming and difficulty in transferring knowledge from single-agent to multi-agent settings. However, a more recent work~\cite{yu2022surprising} show that on-policy methods such as MAPPO (a multi-agent version of proximal policy optimization) outperforms all other SOTA methods including MADDPG and QMIX in various multi-agent benchmarks, and especially works extremely well in complex SMAC settings. \cite{yu2022surprising} show that with proper choices of various hyper-parameter factors including value normalization, clipping ratio, and batch size, etc., policy gradient methods can achieve both data efficiency and high return for agents. Motivated by this promising results of MAPPO, our work focuses on improving the performance of policy gradient methods in MARL. 

We consider a multi-agent environment in which there are agents who attempt to form allies with each other to play against a team of opponents in a partially observable MDP environment where allied agents have to make decision independently without communicating with other members. Unlike the leading method, MAPPO, which focuses on the investigation of hyper-parameter choices for PPO to perform well in the MARL setting, we attempt to enhance the performance of PPO in MARL with the introduction of a novel opponent-behavior-prediction component. This new component is incorporated into the MAPPO framework to enhance the policy learning of allied agents in the MARL setting. A key challenge in our problem setting is that agents in the alliance are unaware of actions taken by their opponents. Not only that, each allied agent only has local observations of opponents locating in the current neighborhood of the agent --- the locations and neighborhoods of all players are not fixed, but instead, are changing over time depending on actions taken by players and the dynamics of the environment. Lastly, learning the behavior of the opponents occurs during the policy learning process of the allied agents. The inter-dependency between these two learning components makes the entire learning process significantly challenging. 

We address the aforementioned challenges while providing the following key contributions. \emph{First}, we convert the problem of opponent behavior learning into predicting next states of the opponents. This way, we don't have to directly predict the behavior of the opponents (of which actions are not observable), but instead, we use the next state prediction outcome as an indirect implication of their behavior. We then cast the problem of opponent next-state prediction as a new multi-agent imitation learning (IL) problem. We propose a new multi-agent IL algorithm, which is an adaptation of IQ-Learn~\cite{garg2021iq} (a SOTA imitation learning algorithm), that only considers local opponent-state-only observations. Especially, instead of imitating opponents' policy, our IL algorithm targets the prediction of next states of the neighboring opponents. \emph{Second}, we provide a comprehensive theoretical analysis which provides bounds on the impact of the changing policy of the allied agents (as a result of the policy learning process) on our imitation learning outcomes. 

\emph{Third}, we present a unified MARL algorithmic framework in which we incorporate our behavior learning component into MAPPO. The idea is that we can combine each allied agent's local observations with the next-state prediction of neighboring opponents of that agent, creating an augmented input based on which to improve the decision making of the allied agent at every state. This novel integration results in a new MARL algorithm, which we name \textit{\textbf{I}mitation-enhanced \textbf{M}ulti-\textbf{A}gent E\textbf{X}tended 
PPO} (IMAX-PPO).

\emph{Finally}, we conduct extensive experiments in several benchmarks ranging from complex to simple ones, including: SMACv2 (an advanced version of the Star-Craft multi-agent challenge)~\cite{ellis2022smacv2}, Google research football (GRF)~\cite{kurach2020google}, and Gold Miner~\cite{Miner}. Our empirical results show that our new algorithm consistently outperforms SOTA algorithms (i.e., QMIX and MAPPO) significantly accross all these benchmarks.
\section{Related Work}
\noindent\textbf{MARL. }
The literature on MARL includes centralized and decentralized algorithms. While centralized algorithms \citep{claus1998dynamics} learn a single joint policy to produce joint actions of all the agents, decentralized learning \citep{littman1994markov} optimizes each agent's local policy independently. There are also algorithms based on \textit{centralized training and decentralized execution} (CTDE). For example, methods in \citep{lowe2017multi,foerster2018counterfactual} adopt actor-critic structures and learn a centralized critic that takes global information as input. Value-decomposition (VD) is a class of methods that represent the joint Q-function as a function of agents’ local Q-functions \citep{sunehag2017value,rashid2020monotonic}.
On the other hand, the use of policy-gradient methods, such as PPO \citep{schulman2017proximal}, has been investigated in multi-agent RL. For example, \cite{de2020independent} propose independent PPO (IPPO), a decentralized MARL, that can achieve high success rates in several hard SMAC maps. IPPO is, however, overall worse than QMIX \citep{rashid2020monotonic}, a VD method. Recently, \citep{yu2022surprising} develop MAPPO, a PPO-based MARL algorithm that outperforms QMIX on some popular multi-agent game environments such as SMAC \citep{samvelyan2019starcraft} and GRF \citep{kurach2020google} environments. {To the best of our knowledge, MAPPO is currently a SOTA algorithm for MARL. In this work, we integrate a new IL-based  behavior prediction model into MAPPO, resulting in a new MARL algorithm that outperforms SOTA MARL methods on various challenging game tasks.}

\noindent\textbf{Imitation Learning (IL). }
In this study, we employ Imitation Learning to anticipate opponents' behavior, making it pertinent to delve into the relevant literature. IL has been acknowledged as a compelling approach for sequential decision-making \citep{ng2000algorithms,abbeel2004apprenticeship}. In IL, a collection of expert trajectories is provided, with the objective of learning a policy that emulates behavior similar to the expert's policy. One of the simplest IL methods is Behavioral Cloning (BC), which aims to maximize the likelihood of the expert's actions under the learned policy. BC disregards environmental dynamics, rendering it suitable only for uncomplicated environments. Several advanced IL techniques, encompassing environmental dynamics, have been proposed \citep{reddy2019sqil,fu2017learning,ho2016generative}. While these methods operate in complex and continuous domains, they involve adversarial learning, making them prone to instability and sensitivity to hyperparameters. The  IQ-learn \citep{garg2021iq} stands as a cutting-edge IL algorithm with distinct advantages, specifically its incorporation of dynamics awareness and non-adversarial training. It's important to note that all the aforementioned IL methods were designed for single-agent RL. In contrast, the literature on multi-agent RL is considerably more limited, with only a handful of studies addressing IL in multi-agent RL. For instance, \citep{song2018multi} presents an adversarial training-based algorithm, named Multi-agent Generative Adversarial IL. It's worth noting that all the IL algorithms mentioned above are established on the premise that actions are observable, which implies that no existing algorithm can be directly applied to our multi-agent game settings with local state-only observations.

\section{Competitive Multi-Agent POMDP Setting}
We consider a multi-player Markov game in which there are multiple agents forming an alliance to play against some opponent agents. We present the Markov game as a tuple
$\{\cS, \cN_\A, \cN_\E, \cA^\A, \cA^\E, P, R \}$, where $\cS$ is the set of global states shared by all the agents, $\cN_\A$ and $\cN_\E$ are the set of ally and enemy agents, $\cA^\A = \prod_{i\in \cN_\cA} \cA^\A_i$ is the set of joint actions of all the ally agents,  $\cA^\E = \prod_{j\in \cN_\E} \cA^\E_j$ is the set of joint actions of all the ally agents, $P$ is the transition dynamics of the game environment, and $R$ is a general reward function that can take inputs as states and actions of the ally or enemy agents and return the corresponding rewards. At each time step where the global state is $S$, each ally agent $i\in \cN_\A$ makes an action $a^{\alpha}_{i}$ according to a policy $\pi^\A_i(a^{\alpha}_{i}|o^{\alpha}_{i})$, where $o^{\alpha}_{i}$ is is the {observation} of ally agent $i$ given state $S$. The joint action can be now defined as follows: 
$$
A^\A = \{a^{\alpha}_{i}|~ i\in \cN_\A\}
$$
and the joint policy is defined accordingly:
$$\Pi^\A(A^\A| S) = \prod\nolimits_{i\in\cN^\A} \pi^\A_i(a^{\alpha}_{i}|o^{\alpha}_{i}).$$ 
The enemy agents, at the same time, make a joint action $A^\E = \{a^\E_{j}|~ j\in \cN_\E\}$ with the following probability:
\[
\Pi^\E(A^\E| S) = \prod\nolimits_{j\in\cN^\E} \pi^\E_j(a^\E_{j}|o^\E_{{j}}).
\]
After the allies and enemies make decisions, the global state transits to a new state $S'$ with the probability $P(S'|A^\E,A^\A,S)$. In our  setting, the enemies' policies $\Pi^\E$ are fixed. Therefore, we can treat the enemies' policies as a part of the environment dynamics, as follows:
\[
P(S'|A^\A,S) =  \sum\nolimits_{A^\E} \Pi(A^\E| S) P(S'| A^\E, A^\A, S)
\]
Our goal is to find a policy that optimizes the allies' expected joint reward, which can be formulated as follows:\footnote{Environment dynamics are implicitly involved in sampling.}
\begin{equation}
    \max\nolimits_{\Pi^\A} \bbE_{(A^\A,S)\sim \Pi^\A}\big[R^\A(S,A^\A)\big] 
\end{equation}
The game dynamics involve both the environment dynamics and the joint policy of enemies, making the training costly to converge. { Our idea is, therefore, to migrate the uncertainties associated with these game dynamics by first predicting the opponent behavior prediction based on past observations of the allies and leveraging this behavior prediction into guiding the policy training for the allies. }
\section{Opponent Behavior Prediction}

The key challenge in our problem setting is that actions taken by opponents are hidden from allied agents. Moreover, each allied agent has limited observations of the other agents in the environment; they can only obtain information about nearby opponents. For example, in the SMAC environment, for each allied agent, besides information about the agent itself, the allied agent is also aware of the relative position, relative distance, and health point, etc. of the neighboring opponents.\footnote{The neighborhood of each allied agent changes over time depending on actions of all agents and the dynamic of the environment.} Therefore, instead of directly predicting opponents' next moves, we focus on anticipating next states of opponents --- this next-state prediction can be used as an implication of what actions have been taken by the neighboring opponents. 

Our key contributions here include: (i) a novel representation of this opponent behavior prediction (or next-state prediction) problem in the form of multi-agent imitation learning; (ii) a new adaptation of IQ-Learn (a SOTA method for solving the standard single-agent imitation learning problem) to solve our new imitation learning problem; (iii) a comprehensive theoretical analysis on the influence of learning policies of the agents in the allies on the next-state prediction outcomes; and (iv) a practical multi-agent imitation learning algorithm which is tailored to local observations of allied agents.  


\subsection{State-Only  Multi-Agent Imitation Learning}
In this section, we first present our approach of modeling the problem of predicting the opponents' behavior as a state-only multi-agent IM. We then describe our adaptation of IQ-Learn for solving our new multi-agent IL problem.\footnote{For the sake of representation, this section focuses on multi-agent IL with \emph{global} state-only observations in which actions of opponents are hidden from the allies. We then introduce a new practical algorithm in next section which localizes this global state-based IL algorithm to adapt with local observations of allied agents.} 

\paragraph{Opponent Behavior Prediction as a Multi-Agent IM.}In order to formulate the problem as a multi-agent IM, we introduce a new notion of a joint reward function for the enemies as $R^\E(S'|A^\A,S)$ associated with a tuple $(S', A^\A,S)$ where $(S, S')$ are the current and next global states and $A^\A$ is the joint action taken by the allies. Intuitively, the next state $S'$ is a joint ``action'' of the enemies in this context. This action ($S'$) is basically a result of the joint actions of the allies $A^\A$ and the \emph{hidden} joint actions of the enemies $A^\E$ in the original multi-agent POMDP. Altogether with the reward function $R^\E(S'|A^\A,S)$, we introduce a new notion of joint policy for the enemies,  $\Pi^\E(S'|S,A^\A)$, which is essentially the probability of ending up at a global state $S'$, from state $S$ when the allies make action $A^\A$. Let $\bPi$ be the support set of the imitating policy, i.e., $\bPi  = \{\Pi:\cS\times\cS\times\cA^\A\rightarrow [0,1],~ \sum_{S'\in \cS} \Pi(S'|S,A^\A) = 1,~ \forall S,S'\in \cS, A^\A\in \cA^\A \}.$ 

We now introduce the maximum-entropy inverse RL framework \citep{ho2016generative} w.r.t the new notions of enemies' reward and policy $(R^\E\!(S'|S,A^\A), \Pi^\E\!(S'|S,A^\A))$ as follows: 
\begin{align}
 & \max\nolimits_{R^\E} \min\nolimits_{\Pi\in \bPi} \Big\{L(R^\E,\Pi) 
  =\bbE_{\rho_E^e}[R^e(S'|S,A^\A)- \nonumber\\ 
  &\qquad\bbE_{\rho_{\Pi}^e}[R^e(S'|S,A^\A)] + \bbE_{\rho_{\Pi}^e}[\ln \Pi(S'|S,A^\A)]\Big\} \label{eq:obj-IRL}
\end{align}
where  $\rho_E^e$ is the occupancy measure of the enemies' joint policy $\Pi^\E$ and $\rho_{\Pi}^e$ the occupancy measure of the imitating policy $\Pi$, and the last term is the entropy regularizer.  

\paragraph{An Adaptation of IQ-Learn.}

Drawing inspiration from the state-of-the-art IL algorithm known as IQ-learn, we construct our IL algorithm which is an adaptation of IQ-Learn tailored to our multi-agent competitive game environment. 

The main idea  of IQ-learn is to convert a reward learning problem into a Q-function learning one. To apply IQ-Learn to our setting, we present the following new \textit{soft and inverse soft} Bellman operators, which is a modification from the original ones introduced in~\cite{garg2021iq}: 
\begin{align}
    \cB^{\Pi,R^\E}_{{Q}^\E}(S'|S,A^\A) &=  R^\E(S'|S,A^\A) + \gamma {V}^\E_\Pi(S')\label{bellman}\\
    \cT^\Pi_{{Q}^\E}(S'|S,A^\A) &= {Q}^\E(S'|S,A^\A) - \gamma {V}^\E_\Pi(S') \label{inverse.bellman}
\end{align}
where
{ \[
{V}^\E_\Pi(S)\!=\!\bbE_{(A^\A,S')\sim (\Pi^\A,{\Pi}) }\!\Big[{Q}^\E(S'|S,A^\A) \!-\! \ln({\Pi}(S'|S,A^\A))\Big] 
\]}
The key difference in Equations~\ref{bellman} and~\ref{inverse.bellman} in comparison with the original ones presented in IQ-Learn is that actions defined for IL in our setting coincide with next states, resulting in no computation of expectation for future returns. This difference may raise an important question of whether properties and theoretical results of original IQ-Learn still hold for our new problem setting. Interestingly, we show that key properties of IQ-Learn still carry out to this new setting. 
 
First, it can be seen that the soft Bellman operator $ \cB^{\Pi,R^\E}_{{Q}^\E}$  is contractive, thus defining a unique fixed point solution ${Q}^*$ such that $\cB^{\Pi,R^e}_{Q^*} = Q^*$.
Let us further define the following function of $\Pi$ and $Q^\E$ for the Q-function learning:
\begin{align}
&{J}(\Pi,{Q}^\E) =  \bbE_{\rho_E^e}[ \cT^\Pi_{{Q}^\E}(S'|S,A^\A)] \nonumber \\
&\qquad\qquad- \bbE_{\rho_{\Pi}^e}[ \cT^\Pi_{{Q}^\E}(S'|S,A^\A)] + \bbE_{\rho_{\Pi}^e}[\ln \Pi(S'|S,A^\A)]\nonumber 
\end{align}
We obtain a similar result as in IQ-Learn showing a connection between the learning reward and learning Q-functions:\footnote{All proofs are in the appendix.}
\begin{proposition}\label{prop:IQ-equip}
    For any reward function $R^e$, let  $Q^*$ be the unique fixed point solution to the soft Bellman equation $ \cB^{\Pi,R^\E}_{Q^*} = Q^*$, then we have
  $L(\Pi,R^\E) = {J}(\Pi,Q^*)$, and for any $Q^\E$,
   ${J}(\Pi,Q^\E) = L(\Pi,\cT^\Pi_{Q^\E})$.
\end{proposition} 
Proposition \eqref{prop:IQ-equip} indicates that  the reward learning problem $\max_{R^\E}\min_{\Pi} L(\Pi,R^\E)$ can be converted equivalently into the Q-learning one $\max_{Q^\E}\min_{\Pi} J(\Pi,Q^\E)$. Suppose $Q^*$ is a solution to the Q-learning problem, then rewards can be recovered by taking $R^e(S'|S,A^\A) = Q^*(S'|S,A^\A) - \gamma V^e_\Pi(S')$. Furthermore, our following Proposition~\ref{J.compact} shows that the function  ${J}(\Pi,Q^\E)$ can be reformulated in a more compact form that is convenient for training.
\begin{proposition}\label{J.compact}
    The  function $J(\cdot)$ can be written as follows:
    \begin{align}
        J(\Pi,Q^e)  &=  \bbE_{(S',A^\A,S)\sim \rho^\E_E}\Big[{Q}^\E(S'|S,A^\A) - \gamma  {V}^e_\Pi(S')\Big] \nonumber\\
        &\qquad\qquad\qquad+ (1-\gamma) \bbE_{S^0\sim P^0}{V}^\E_\Pi(S^0)\label{eq:J.compact}
    \end{align}
\end{proposition}
\noindent where $S_0$ is an initial state. Under this viewpoint, we show that key properties of classical IQ-learn still hold in the context of our multi-agent with missing observations (Proposition~\ref{th:th1}), making our IQ-learn version convenient to use.  

\begin{proposition}
\label{th:th1}
   The problem   $\max_{Q^\E} \min_{\Pi^\E} J(\Pi^\E,Q^\E) $ is equivalent to the maximization 
$
{\max_{Q^\E}}J(\Pi^Q,Q^\E) 
$
where 
$$\Pi^Q(S'|S,A^\A) = \frac{\exp(Q^\E(S'|S,A^\A))}{ \sum_{S^{''}}\exp(Q^\E(S^{''}|S,A^\A))}
$$ 
Moreover, the function $J(\Pi^Q,Q^\E)$ is concave in $Q^\E$.     
\end{proposition}

\begin{corollary}\label{coro:1}
    If $\Pi = \Pi^Q$ (defined in Theorem \ref{th:th1}), then we can write  $V^\E (S)$ as follows:
{\small\begin{align*}
V^\E_\Pi (S) &\!=\! V^\E_Q (S) \!=\! \sum\nolimits_{A^\A} \!\!\!\Pi^\A(A^\A|S) \ln\left(\!\sum\nolimits_{S'}\!\!\exp\!\left(Q^\E(S'|S,A^\A)\right)\!\right)
\end{align*}}
    
\end{corollary}

\subsection{Affect of Allies' Policies on Imitation Learning}

The above IL algorithm is trained during the training of the allies' policies, thus the game dynamics change during the IL process. We analyze the impact of these changes on the imitation policy. To this end, we consider the loss function of the imitation model as a function of the allies' joint policy.
\begin{align}
\Phi(\Pi^\A|Q^\E)  &\!=\!  \bbE_{(S',A^\A,S)\sim (\Pi^{\E},\Pi^\A)}\!\! \left[Q^\E(S'|S,A^\A) \!-\! \gamma  V^\E_Q(S')\right]\nonumber\\
&~~~~~~~~~~~~~~~~~~-(1-\gamma) \bbE_{S^0\sim P^0}V^\E_Q(S^0) \nonumber
\end{align}
In the following, we first present our theoretical results about bounds on the impact of the allies' changing policies on the imitation learning's loss function (Proposition~\ref{prop:p1} for a discrete state space and Proposition~\ref{prop:bound-continous} for a continuous space). Based on these results, we then provide a theoretical bound on the IL learning outcome accordingly (Corollary~\ref{coro.1}). 
\begin{proposition} [Finite state space $\cS$]
\label{prop:p1}
Given two allies' joint policies $\Pi^\A$ and $\widetilde{\Pi}^\A$ such that $\KL(\Pi^\A(\cdot|S) ||\widetilde{\Pi}^\A(\cdot|S))\leq \epsilon$ for any state $S\in \cS$, the following inequality holds:
{\small\begin{align}
&\left|\Phi(\Pi^\A|Q^\E) -  
\Phi(\widetilde{\Pi}^\A|Q^\E)\right| \nonumber\\
&\qquad\leq  \left(\frac{\gamma+1+(1-\gamma)^3}{(1-\gamma)^2}\overline{Q} +\frac{1+(1-\gamma)^3}{(1-\gamma)^2} \ln |\cS| \right)\sqrt{2\ln 2 \epsilon  }\nonumber
\end{align}}
where $\overline{Q} \!=\! \!\max\limits_{(S'_{t},S_{t},A^\A_t)}\!\!Q^\E(S'_{t}|S_{t},A^\A_t)$ is an upper bound of $Q^\E$. 
\end{proposition}
Under a continuous state space, an actor-critic method is used as the policy $\Pi^Q$ cannot be computed exactly. In this case, one can use an explicit policy $\Pi$ to approximate $\Pi^Q$. We then iteratively update $Q^\E$ and $\Pi$ alternatively using the loss function in Proposition \ref{J.compact}. 
In particular, for a fixed $Q^\E$, a soft actor update $\max_{\Pi}\bbE_{S' \sim \Pi(S|A^\A, S)}[Q^\E(S'|S,A^\A)-\ln \Pi(S'|S,A^\A)]$ will bring $\Pi$ closer to $\Pi^Q$. 

Let $\Phi(\Pi^\A|\Pi,Q^\E)$ be the objective of IL in \eqref{eq:J.compact}, written as a function of the allies' joint policy $\widetilde{\Pi}^\A$. 
The following  proposition establishes a bound for the variation of $|\Phi(\Pi^\A|\Pi,Q^\E) - \Phi(\widetilde{\Pi}^\A|\Pi,Q^\E)|$ as funntion of $\KL(\Pi^\A||\widetilde{\Pi}^\A)$.
\begin{proposition}[Continuous state space $\cS$]\label{prop:bound-continous}
Given two allies' joint policies $\Pi^\A$ and $\widetilde{\Pi}^\A$ such that for every state $S\in \cS$, $\KL(\Pi^\A(\cdot|S) ||\widetilde{\Pi}^\A(\cdot|S))\leq \epsilon$, the following inequality holds:
{\small\begin{align}
&\left|\Phi(\Pi^\A|Q^\E,\Pi) -  
\Phi(\widetilde{\Pi}^\A|Q^\E ,\Pi)\right| \nonumber\\
&\qquad\leq  \left(\frac{\gamma+1+(1-\gamma)^3}{(1-\gamma)^2}\overline{Q}  - \frac{1+(1-\gamma)^3}{(1-\gamma)^2} H \right)\sqrt{2\ln 2 \epsilon  }\nonumber
\end{align}}
where $\overline{Q}=\sup\limits_{(S'_{t},S_{t},A^\A_t)}\!\!Q^\E(S'_{t}|S_{t},A^\A_t)$ is an upper bound of $Q^\E(\cdot|\cdot,\cdot)$ and $H = \inf\limits_{(S,A^\A)}\sum_{S}\Pi(S'|S,A^\A)\ln \Pi(S'|S,A^\A)$, is a lower bound of the entropy of the actor policy $\Pi(\cdot|\cdot,\cdot)$.    
\end{proposition}

Propositions~\ref{prop:p1} \& \ref{prop:bound-continous} allow us to establish an upper bound for the imitation learning when 
the allies' joint policy changes.
{\begin{corollary}\label{coro.1}
Given two allies' joint policies $\Pi^\A$ and $\widetilde{\Pi}^\A$ with $\forall S\!\in\! \cS$, $\KL(\Pi^\A(\cdot|S) ||\widetilde{\Pi}^\A(\cdot|S)\!\leq\! \epsilon$, the following holds:
{\small\begin{align*}
&\Big|\max\nolimits_{Q^\E}\!\!\big\{\Phi(\Pi^\A|Q^\E)\big\} -  
\max\nolimits_{Q^\E}\!\!\big\{\Phi(\widetilde{\Pi}^\A|Q^\E)\big\}\Big| \leq \cO(\sqrt{\epsilon}) \;\text{ (discrete)}\\   
&\Big|\max\nolimits_{Q^\E}\min\nolimits_\Pi\big\{\Phi(\Pi^\A|Q^\E,\Pi)\big\}\qquad\qquad\qquad\qquad\quad\text{ (continuous)}\nonumber\\
&\qquad\qquad\quad-  
\max\nolimits_{Q^\E} \min\nolimits_{\Pi}\big\{\Phi(\widetilde{\Pi}^\A|Q^\E,\Pi)\big\}\Big| \leq \cO(\sqrt{\epsilon}). 
\end{align*}}
\end{corollary}}
Since the allies' joint policy $\Pi^\A$  will be changing during our
policy learning process, the above result implies that the imitating
policy will be stable if $\Pi^\A$ becomes stable, and if $\Pi^\A$ is converging to a target policy $\Pi^{\A*}$, then the imitator’s policy also converges to the one that is trained with the target ally policy with a rate of $\sqrt{\KL (\Pi^\A||\widetilde{\Pi}^\A)}$.
That is, if the actual policy is within a $\cO(\epsilon)$ neighborhood of  the target policy (i.e., $\KL (\Pi^\A||\widetilde{\Pi}^\A)\leq \epsilon$)
then the expected return of the imitating policy is within a $\cO(\sqrt{\epsilon})$
neighborhood of the desired ``\textit{expected return}'' given by the target policy.

\begin{figure*}[t!]
\centering
\includegraphics[width=1.0\linewidth]{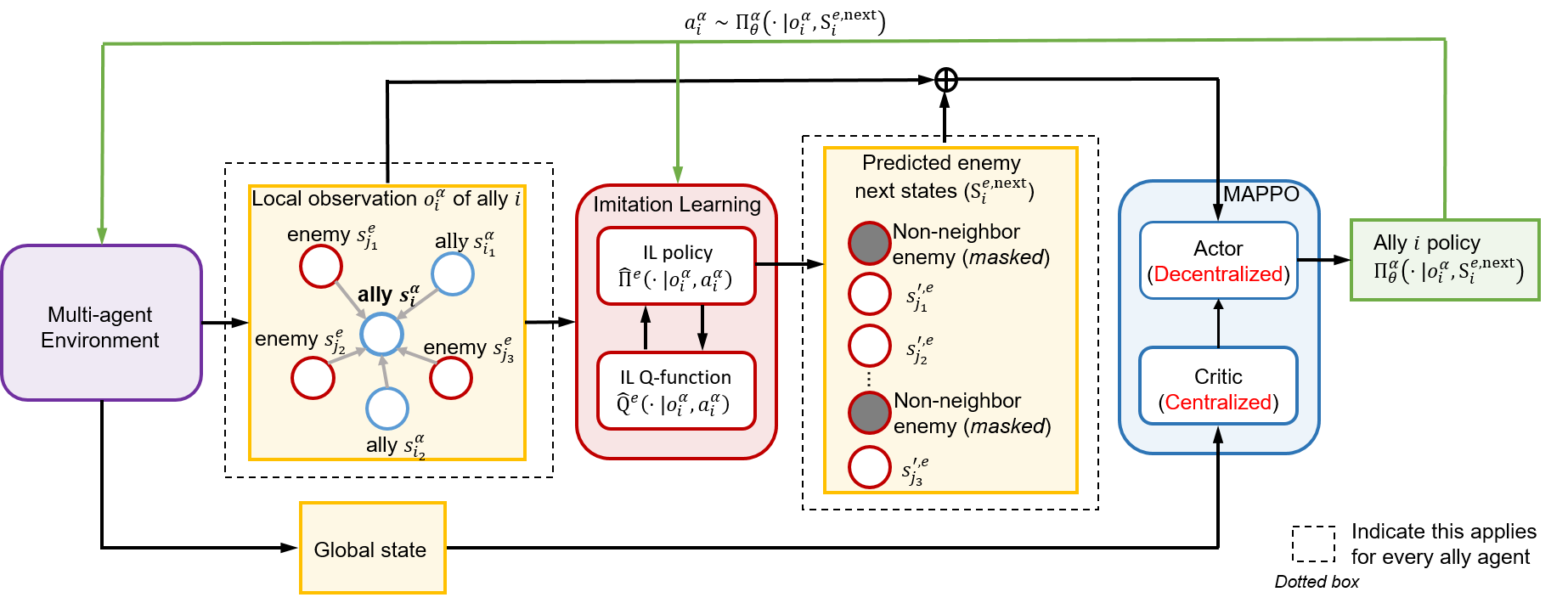}
\caption{An overview of our IMAX-PPO algorithm. Each local observation $o^{\alpha}_i$ of an ally agent $i$ includes information about itself, as well as enemy and ally agents in its neighborhood (which changes over time). The output of the IL component is the predicted next states of neighboring enemy agents (predictions for the non-neighbor enemies will be masked out). }
\label{fig:IMAX}
\end{figure*}
\section{IMAX-PPO: \textbf{I}mitation-enhanced \textbf{M}ulti-\textbf{A}gent E\textbf{X}tended 
PPO Algorithm }
We present our MARL algorithm for the competive game setting. We first focus on a practical implementation of an imitation learning algorithm taking into account local observations. We then show how to integrate this into our MARL algorithm. We call our algorithm as IMAX-PPO, standing for \textit{\textbf{I}mitation-enhanced \textbf{M}ulti-\textbf{A}gent E\textbf{X}tended 
PPO} algorithm.

\subsection{Imitation Learning with Local Observations}
In previous section, we present our new imitation learning algorithm (which is an adaptation of IQ-Learn) to learn an enemy policy $\widehat{\Pi}^\E(S'|S,A^\A)$ that behaves similarly to the probabilities of ending up at state $S'$ when the current global state is $S$  and the allies' joint action is $A^\A$. Here, $S$ represents all the information that the enemy agents can access.
In our multi-agent game setting, however, ally agents do not have access to the observations of the enemies. Instead, each ally agent $i$ only has information about its neighboring enemies observable by agent $i$ (such as their locations, speeds, etc.). 

Therefore, we adapt our IL algorithm in accordance with such available local information. That is, the aim now is to optimize policies $\widehat{\Pi}^\E_{\psi_\pi}(S^{\E,\text{next}}_i|o^\A_i, a^\A_i)$, $i\in \cN^\A$, where $o^\A_i$ is an observation vector of agent $i$,
containing the local states of the agent $i$ itself as well as of all the agents in the neighborhood that are observable by agent $i$. In particular,
$$o^{\alpha}_i = \{s^{\alpha}_i\}\cup \{s^{\alpha}_{i_k}\!: i_k\in N(i) \cap \cN^{\alpha}\} \cup \{s^{\E}_{j_k}\!: j_k\in N(i) \cap \cN^{\E}\}$$
where $s^{\alpha}_{i_k}$ is the local state of an ally agent $i_k$ and $s^{\E}_{j_k}$ is of an enemy agent $j_k$ and $N(i)$ is the neighborhood of $i$. 

To apply our multi-agent IL to this local observation setting, we build a common policy network  $\widehat{\Pi}^\E_{\psi_\pi}$  and Q network $\widehat{Q}^\E_{\psi_Q}$ for all the agents where $\psi_\pi$ and $\psi_Q$ are the network parameters. The objective function for IL can be reformulated according to local observations of the allies as follows: 
  \begin{align}
        &J(\widehat{\Pi}^\E_{\psi_\pi},\widehat{Q}^\E_{\psi_Q})  =  \!\!\sum\nolimits_{i\in \cN^\A}\!\!\bbE_{(S^{\E,\text{next}}_i,a^\A_i,o^\A_i)\sim \rho^\E_E}\!\Big[\widehat{Q}^\E\!(S^{\E,\text{next}}_i|o^\A_i,a^\A_i)\nonumber\\
        &\qquad\quad- \gamma  {V}^\E_\Pi(S^{\E,\text{next}}_i)\Big]        + (1-\gamma) \bbE_{S^{\E0}_i\sim P^0}{V}^\E_\Pi(S^{\E0}_i)\label{eq:J.compact.local}
    \end{align}
where $S^{\E,\text{next}}_i= \{s'^{\E}_{j_k}\!:\! j_k \!\in \!N(i)\cap \cN^\E\}$ is the next states of enemies in the current neighborhood of ally agent $i$. In addition, ${V}^\E_\Pi(S^{\E,\text{next}}_i) =  \bbE_{(a^\A_i,S^{\E,\text{next}}_i)\sim (\Pi^\A,{\widehat{\Pi}^\E_{\psi_\pi}}) }\!\big[\widehat{Q}^\E_{\psi_Q}(S^{\E,\text{next}}_i|o^\A_i,a^\A_i) \!-\! \ln(\widehat{\Pi}^\E_{\psi_\pi}(S^{\E,\text{next}}_i|o^\A_i,a^\A_i))\big]$. We can update $\widehat{\Pi}^\E_{\psi_\pi}$  and  $\widehat{Q}^\E_{\psi_Q}$ by the following actor-critic rule: for a fixed $\widehat{Q}^\E_{\psi_Q}$, update $\psi_Q$ to maximize $J(\widehat{\Pi}^\E_{\psi_\pi},\widehat{Q}^\E_{\psi_Q})$, and for a fixed $\widehat{\Pi}^\E_{\psi_\pi}$, apply soft actor-critic (SAC) to update $\psi_\pi$. 


\subsection{IMAX-PPO Algorithm}
We aim to optimize the policy $\Pi^\A_\theta(a^\A_i|o^\A_i,S^{\E,\text{next}}_i), i\!\in\! \cN^\A\}$ of the allies that optimizes the long-term expected joint reward:  
\[
\max\nolimits_{\Pi^{\A}_\theta}\bbE_{(a^\A_i,o^\A_i,S^{\E,\text{next}}_i)\sim \Pi^{\A}_\theta } \Big[\sum\nolimits_{i\in\cN^\A} R^\A_i(a^\A_i,o^\A_i)\Big]
\]
where $o^\A_i$ is an observation vector of agent $i$, $S^{\E,\text{next}}_i$  is the information derived from the imitator for agent $i$,  $a^\A_i \in \cA^\A_i$ is a local action of agent $i$. To facilitate the training and integration of the imitation learning policy into the MARL algorithm, for every ally agent $i$, we gather game trajectories following the structure $(o_{i}, a^\A_{i}, S^{\E,\text{next}}_{i})$.  These gathered observations are then stored in a replay buffer to train the imitation policy $\widehat{\Pi}^\E_{\psi_\pi}(S^{\E,\text{next}}_i|o^\A_i,a^\A_i)$.

To integrate the imitator into the IMAX-PPO framework, we include the predicted next states of the enemy agents as inputs for training the allies' policy $\Pi^\A$. Specifically, at each game state $S$, considering the current actor policy $\Pi^\A$ and the imitating policy $\widehat{\Pi}^\E$, for each agent $i\in \cN^\A$, we draw a sample for the allied agents' joint action $\widetilde{A}^\A \!\sim\! \Pi^\A$. Corresponding local observation $o^\A_i$ and action $\widetilde{a}^\A_i$ of each agent $i$ are then fed as inputs into the imitation policy to predict the subsequent state $S^{\E,\text{next}}_{i} \!\sim\! \widehat{\Pi}^\E(\cdot|o^\A_{i},\widetilde{a}^\A_{i})$. Once the predicted local states $\{S^{\E,\text{next}}_i,~ i\!\in \!\cN^\A\}$ are available, it is used as input to the actor policy $\Pi^\A_\theta$ in order to generate new actions for the allied agents. In simpler terms, we select a next local action $a'^{\A}_{i} \!\sim\! \Pi^\A_\theta (\cdot|o^{\alpha}_{i}, S^{\E,\text{next}}_{i})$.
{Beside the allies' policy network, we also use a centralized value network $V^\A_{\theta_v}(S)$ and update it together with the policy network in an actor-critic manner,  similarly to  MAPPO}.
The actor-network is trained by optimizing the following objective:
\begin{align}
L^\A(\theta) &= \sum\nolimits_{i\in \cN^\A}\bbE_{o^\A_i,a^\A_i,S^{\E,\text{next}}_i}\big[\min\{r_i(\theta) \widehat{A},\nonumber \\
&\qquad\qquad\qquad\qquad\text{clip}(r_i(\theta),1-\epsilon,1+\epsilon)\widehat{A}\}\big]\label{eq:MAPPO-actor}    
\end{align}
where $r_i(\theta) = \frac{\Pi^\A_\theta(a^\A_i|o^\A_i,S^{\E,\text{next}}_i)}{\Pi^\A_{\theta_{old}}(a^\A_i|o^\A_i,S^{\E,\text{next}}_i)}$
and $\widehat{A}$ is the advantage function, calculated by  Generalized Advantage Estimation (GAE). The Critic network is trained by optimizing
\begin{align*}
\Phi^\A\!(\theta_v) \!=\! \bbE_{S}\!\Big[\!\max\{(V^\A_{\theta_v}(S) \!-\! \widehat{R}(S))^2\!,(V^{\text{clip}}_{\theta_v,\theta_{v,old}}\!(S)\!-\!\widehat{R}(S))^2\}\!\Big] 
\end{align*}
where $\widehat{R}(S) = \widehat{A}+ V^\A_{\theta_{v,old}}(S)$ and $V^{\text{clip}}_{\theta_v,\theta_{v,old}}(S) = \text{clip}(V^\A_{\theta_v}(S),V^\A_{\theta_{v,old}}(S)-\epsilon, V^\A_{\theta_{v,old}}(S)+\epsilon)$.
We provide the key stages in Algorithm \ref{alg:IMAX} below. Additionally, Fig. \ref{fig:IMAX} serves as an illustration of our IMAX-PPO algorithm.

\begin{algorithm}[t!]
\caption{IMAX-PPO Algorithm}
\label{alg:IMAX}
\textbf{Input}: Initial allies' policy network $\Pi^\A_\theta$, initial allies' value network $V^\A_{\theta_v}$, initial imitator's policy network $\widehat{\Pi}^\E_{\psi_\pi}$, initial imitator's Q network ${Q}^\E_{\psi_Q}$, learning rates $\kappa^\E_\pi,\kappa^\E_Q, \kappa^\A_\pi, \kappa^\A_V$.\\
\textbf{Output}: Trained allies' policy $\Pi^\A_\theta$
\begin{algorithmic}[1] 
\FOR{$ t = 0,1,\ldots$} 
    \STATE \mbox{{\cment{\textbf{\# Updating imitator:}}}}
        \STATE \cment{\# Train Q function using the objective in \eqref{eq:J.compact.local}}
        \STATE $\psi_{Q,t+1} = \psi_{Q,t} + \kappa^\E_Q\nabla_{\psi_Q}[J(\psi_Q)]$
        \STATE \cment{Update policy $\widehat{\Pi}^\E_{\psi_\pi}$ with a SAC style actor update (for continuous domains)}
        \STATE $\psi_{\pi,t+1} = \psi_{\pi,t} - \kappa^\E_{\pi}\nabla_{\psi_{\pi}}\bbE_{S^{\E,\text{next}}_i}[{V}^\E_\Pi(S^{\E,\text{next}}_i)]$
    \STATE \mbox{{\cment{\# \textbf{Updating allies' policy:}}}}
     \STATE \cment{\# Update allies' actor by maximizing $L^\A(\theta)$}   
     \STATE  $\theta_{t+1} = \theta_t + \kappa^\A_\pi \nabla_{\theta} L^\A(\theta)$
     \STATE \cment{\# Update allies' critic by minimizing $\Phi^\A(\theta_v)$}   
     \STATE  $\theta_{v,t+1} = \theta_{v,t} - \kappa^\A_V \nabla_{\theta_v} \Phi^\A(\theta_v)$
\ENDFOR
\STATE \textbf{return} solution
\end{algorithmic}
\end{algorithm}

\section{Experiments}
We evaluate the performance of our algorithm IMAX-PPO in comparison with two baselines which are the two SOTA multi-agent RL algorithms: MAPPO and QMIX. We run extensive experiments in three competitive environments: SMACv2, Google Research Football (GRF), and Miner. Each reported value is computed based on $32$ different rounds of game playing (each corresponds to a different random seed).

\paragraph{SMACv2.}
SMACv2 \citep{ellis2022smacv2} is an advanced variant of SMAC, driven by the aim to present a more challenging setting for the assessment of cooperative multi-agent reinforcement learning algorithms. In SMACv2, scenarios are procedurally generated, which require agents to generalize to previously unseen settings (from the same distribution) during evaluation. This benchmark consists of $15$ sub-tasks where the number of agents varies from $5$ to $20$. The agents can play with opponents of different difficulty levels. 
In comparison to SMACv1~\cite{samvelyan2019starcraft}, SMACv2 stands apart by permitting randomized team compositions, varied starting positions, and an emphasis on augmenting diversity. 

\begin{figure*}[t!]
\centering
\showlegend[0.45]{20.2}{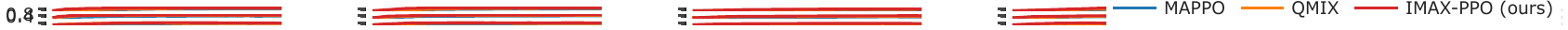} \\

\showinstance[0.2]{0}{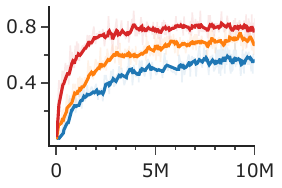}{protoss_5_vs_5}
\showinstance[0.18]{18}{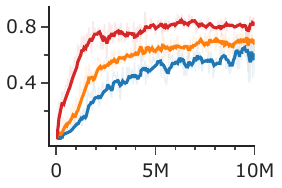}{protoss_10_vs_10}
\showinstance[0.18]{18}{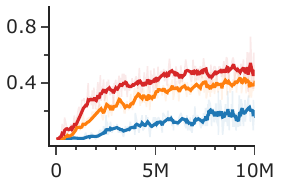}{protoss_10_vs_11}
\showinstance[0.18]{18}{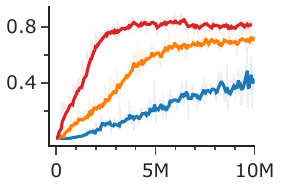}{protoss_20_vs_20}
\showinstance[0.18]{18}{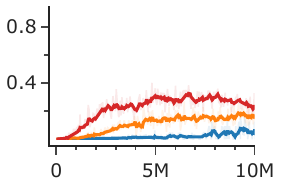}{protoss_20_vs_23}

\showinstance[0.2]{0}{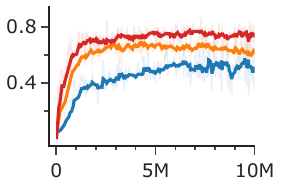}{terran_5_vs_5}
\showinstance[0.18]{18}{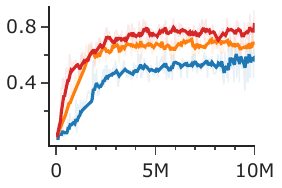}{terran_10_vs_10}
\showinstance[0.18]{18}{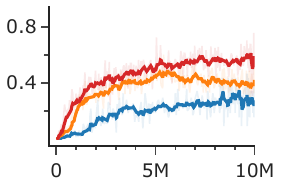}{terran_10_vs_11}
\showinstance[0.18]{18}{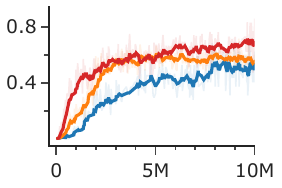}{terran_20_vs_20}
\showinstance[0.18]{18}{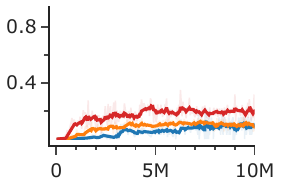}{terran_20_vs_23}

\showinstance[0.2]{0}{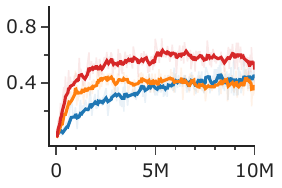}{zerg_5_vs_5}
\showinstance[0.18]{18}{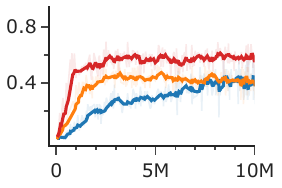}{zerg_10_vs_10}
\showinstance[0.18]{18}{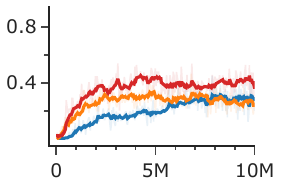}{zerg_10_vs_11}
\showinstance[0.18]{18}{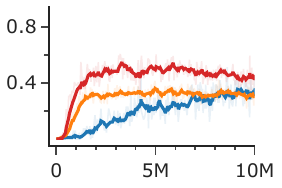}{zerg_20_vs_20}
\showinstance[0.18]{18}{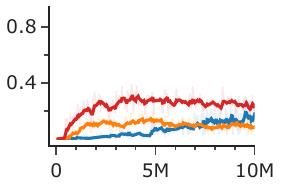}{zerg_20_vs_23}
\caption{Win-rate curves on SMACv2 environment.}
\label{fig:smacv2}
\end{figure*}

Figure \ref{fig:smacv2} shows the performance of the three algorithms during the training process across $15$ sub-tasks. The x-axis is the number of training steps and the y-axis represents the winning rates averaged over $32$ rounds of evaluations. In Figure~\ref{fig:smacv2}, IMAX-PPO consistently and significantly outperforms MAPPO and QMIX. Notably, IMAX-PPO frequently attains quicker convergence --- it achieves high win rates at earlier training stages. This could be attributed to the incorporation of the IL component, which facilitates faster comprehension of opponents throughout the game.
 Finally, details of the win rates at the end of training are shown in Tables~(\ref{tab.protoss}--\ref{tab.zerg}). 

 \begin{table}[t!]
\centering
\begin{tabular}{c|c|c|c}
\hline
Scenarios            & MAPPO   & QMIX    & IMAX-PPO \\ 
\hline
protoss\_5\_vs\_5   & 56.15\% & 69.30\% & \textbf{78.52\%} \\
protoss\_10\_vs\_10 & 56.81\% & 69.58\% & \textbf{82.27\%} \\
protoss\_10\_vs\_11 & 15.33\% & 41.75\% & \textbf{49.44\%} \\
protoss\_20\_vs\_20 & 39.66\% & 71.78\% & \textbf{82.17\%} \\
protoss\_20\_vs\_23 & 4.42\%  & 15.40\% & \textbf{22.55\%} \\
\hline
Average             & 34.48\% & 53.56\% & \textbf{62.99\%} \\ 
\hline
\end{tabular}
\caption{Win-rates on SMACv2's Protoss.}
\label{tab.protoss}
\end{table}

\begin{table}[t!]
\centering
\begin{tabular}{c|c|c|c}
\hline
Scenarios            & MAPPO   & QMIX    & IMAX-PPO \\ 
\hline
terran\_5\_vs\_5   & 47.86\% & 61.33\% & \textbf{75.67\%} \\
terran\_10\_vs\_10 & 56.94\% & 67.47\% & \textbf{82.97\%} \\
terran\_10\_vs\_11 & 26.24\% & 41.39\% & \textbf{57.11\%} \\
terran\_20\_vs\_20 & 53.71\% & 55.95\% & \textbf{70.16\%} \\
terran\_20\_vs\_23 & 9.31\%  & 8.62\%  & \textbf{19.48\%} \\
\hline
Average             & 38.81\% & 46.95\% & \textbf{61.08\%} \\ 
\hline
\end{tabular}
\caption{Win-rates on SMACv2's Terran.}
\label{tab.terran}
\end{table}

\begin{table}[t!]
\centering
\begin{tabular}{c|c|c|c}
\hline
Scenarios            & MAPPO   & QMIX    & IMAX-PPO \\ 
\hline
zerg\_5\_vs\_5   & 45.25\% & 37.22\% & \textbf{49.78\%} \\
zerg\_10\_vs\_10 & 37.93\% & 37.94\% & \textbf{55.62\%} \\
zerg\_10\_vs\_11 & 27.27\% & 24.58\% & \textbf{36.90\%} \\
zerg\_20\_vs\_20 & 34.81\% & 30.58\% & \textbf{43.03\%} \\
zerg\_20\_vs\_23 & 18.30\% & 9.64\%  & \textbf{22.07\%} \\
\hline
Average          & 32.71\% & 27.99\% & \textbf{41.48\%} \\ 
\hline
\end{tabular}
\caption{Win-rates on SMACv2's Zerg.}
\label{tab.zerg}
\end{table}

\paragraph{Google Research Football (GRF).}
This is a challenge on Kaggle competitions made by Google Research team \citep{kurach2020google}. 
We focus on three main sub-tasks, sorted based on increasing difficulty levels: (i) {\textit{academy-3-vs-1-with-keeper}}: three allies try to score  against a goal-keeping opponent; (ii) \textit{{academy-counterattack-easy}}: four allies versus a counter-attack opponent and a goal-keeping opponent; and (iii) \textit{{academy-counterattack-hard}}: four allies versus two counter-attack opponents and a goalkeeper. 

 By default, the representations of all agents' observations are RGB pixels in GRF, so we pre-process this information by distilling some important features such as object positions, object directions, distances between objects, etc. The final win rates are reported in Table \ref{tab:GRF-win-rates}, which shows that IMAX-PPO achieves nearly 100\% win-rates, and significantly outperforms both MAPPO and QMIX.\footnote{Due to limited space, the win rate curves during the training for GRF and Gold Miner environments are in the appendix.}


\begin{table}[htb]
\centering
\begin{tabular}{c|c|c|c}
\hline
Scenarios            & MAPPO   & QMIX    & IMAX-PPO \\ 
\hline
academy\_3\_vs\_1 & 88.03\% & 8.12\%  & \textbf{99.98\%} \\
counterattack\_easy    & 87.76\% & 15.98\% & \textbf{96.52\%} \\
counterattack\_hard    & 77.38\% & 3.22\%  & \textbf{99.64\%} \\
\hline
Average & 84.39\% & 9.11\%  & \textbf{98.71\%} \\ 
\hline
\end{tabular}
\caption{Win-rates on GRF environment}
\label{tab:GRF-win-rates}
\end{table}

\paragraph{Gold Miner.}

This is another competitive multi-agent game  for evaluating our methods. It originates from a multi-agent RL competition.\footnote{Link: \url{https://github.com/xphongvn/rlcomp2020}} Multiple miners navigate in a 2D terrain containing obstacles and repositories of gold. Players get points according to the volume of gold they successfully extract. 
This game is challenging to win as the agents have to learn to play against extremely well-designed heuristic-based enemies. 
In this game,  the ally agents win if the allied team's average mined gold is higher than that of the enemy team.

We customized the original environment into three sub-tasks (between two allies against two enemies) of three difficulty levels: (i) \textit{Easy (easy\_2\_vs\_2)}: The enemies' greedy strategy is to find the shortest way to the golds; (ii) \textit{Medium (medium\_2\_vs\_2)}: One enemy is greedy, and the other follows the algorithm of the second-ranking team in the competition; and (iii) \textit{Hard (hard\_2\_vs\_2)}: The enemies are the first- and second-ranking teams in the competition.

For this environment, the win rates are shown in Figure \ref{fig:miner-win-rates}. Again, IMAX-PPO consistently demonstrated superior win rates across all three tasks. Especially, in the hard-level task, IMAX-PPO manages to win more than $50\%$ of the time against the first and second-ranking teams in the competition. 


\begin{table}[t!]
\centering
\begin{tabular}{c|c|c|c}
\hline
Scenarios            & MAPPO   & QMIX    & IMAX-PPO \\ 
\hline
easy\_2\_vs\_2   & 50.85\% & 56.65\% & \textbf{61.52\%} \\
medium\_2\_vs\_2 & 42.07\% & 47.85\% & \textbf{55.80\%} \\
hard\_2\_vs\_2   & 35.30\% & 41.24\% & \textbf{50.86\%} \\
\hline
Average                 & 42.74\% & 48.58\% & \textbf{56.06\%} \\ 
\hline
\end{tabular}
\caption{Win-rates on Gold Miner environment}
\label{fig:miner-win-rates}
\end{table}

\section{Conclusion}
We introduced a novel principled framework for enhancing agent training in competitive multi-agent environments through imitation learning. Our innovative IL model, adapted from IQ-learn, allows us to mimic opponents' behavior using only partial local state observations. Through the integration of our IL model into a multi-agent PPO algorithm, our IMAX-PPO algorithm consistently outperforms previous SOTA MARL algorithms such as QMIX and MAPPO. This improvement is observed across various challenging multi-agent tasks, including SMACv2 and GRF.
One possible limitation of our approach is its reliance on the assumption that the enemies do not update their policies during training. If this assumption is violated, the IL process could become unstable and exhibit slow convergence. A future direction would be to develop an adaptive multi-agent IL algorithm to make the learning stable and efficient in such settings.


\bibliography{reference}

\newpage
\onecolumn
\appendix
\onecolumn
\section{Missing Proofs}
\subsection{Proof of Proposition \ref{prop:IQ-equip}}
\textbf{Proposition \ref{prop:IQ-equip}:}
\textit{    For any reward function $R^e$, let  $Q^*$ be the unique fixed point solution to the soft Bellman equation $ \cB^{\Pi,R^\E}_{Q^*} = Q^*$, then we have
  $L(\Pi,R^\E) = {J}(\Pi,Q^*)$, and for any $Q^\E$,
   ${J}(\Pi,Q^\E) = L(\Pi,\cT^\Pi_{Q^\E})$.
}

\begin{proof}
       The proof is easily verified, as we can see that if $Q^*$ is a solution to the soft Bellman equation, then $$Q^*(S'|S,A^\A) =  R^e(S'|S,A^\A) + \gamma V^*\Pi(S')$$ where 
    \[
    {V}^{*}_{\Pi}(S)=\bbE_{(A^\A,S')\sim (\Pi^\A,{\Pi}) }\!\!\left[{Q}^*(S'|S,A^\A) \!-\! \ln({\Pi}(S'|S,A^\A))\right]
    \]
    which implies 
    \[
    R^e(S'|S,A^\A) = Q^*(S'|S,A^\A) - \gamma V^*_\Pi(S') = \cT^\Pi_{Q^*}(S'|S,A^\A)
    \]
    which validates $L(\Pi,R^\E) = {J}(\Pi,Q^*)$. The second inequality is just a trivial result from the definition of $J$ and $L$.
\end{proof}

\subsection{Proof of Proposition \ref{J.compact}}
\textbf{Proposition \ref{J.compact}:}
  \textit{  The  function $J(\cdot)$ can be written as follows:
    \begin{align}
        J(\Pi,Q^e)  &=  \bbE_{(S',A^\A,S)\sim \rho^\E_E}\Big[{Q}^\E(S'|S,A^\A) - \gamma  {V}^e_\Pi(S')\Big] + (1-\gamma) \bbE_{S^0\sim P^0}{V}^\E_\Pi(S^0)\label{eq:J.compact}
    \end{align}}
\begin{proof}
  From the definition of $J(\Pi,Q^\E)$ , we write function $J$ as
    \begin{align}
J(\Pi,Q^\E) &=  \bbE_{\rho_E^e}[ \cT^\Pi_{Q^\E}(S'|S,A^\A)]- \bbE_{\rho_{\Pi}^e}[ \cT^\Pi_{Q^\E}(S'|S,A^\A)] - \bbE_{\rho_{\Pi}^e}[\ln \Pi(S'|S,A^\A)]\nonumber \\
&= \bbE_{(S',A^\A,S)\sim \rho^\E_E}\!\! \left[{Q}(S'|S,A^\A) \!-\! \gamma  {V}_\Pi(S')\right] - \bbE_{(S',A^\A,S)\sim \rho^\E_\Pi}\!\! \left[{Q}(S'|S,A^\A) \!-\! \gamma  {V}_\Pi(S')\right] \nonumber\\
&\qquad\qquad - \bbE_{\rho_{\Pi}^e}[\ln \Pi(S'|S,A^\A)]\label{eq:123}
\end{align}
We consider the second  and third terms of \eqref{eq:123}  and write
\begin{align}
    &\bbE_{(S',A^\A,S)\sim \rho^\E_\Pi}\!\! \left[Q^{\E}(S'|S,A^\A) \!-\! \gamma  {V}_\Pi(S')\right]  + \bbE_{\rho_{\Pi}^e}[\ln \Pi(S'|S,A^\A)] \nonumber\\
    &=(1-\gamma)\Bigg(\bbE_{S_t,A^\A_t,S_{t+1} \sim\Pi}\left[\sum_{t=0} \gamma^t Q^{\E}(S_{t+1}|S_t,A^\A_t)\right] - \gamma \left(\bbE_{S_t\sim\Pi}\left[\sum_{t=0} \gamma^{t}V^{\E}_\Pi(S_{t+1}) \right]\right)\nonumber\\
    &\qquad\qquad\qquad+ \bbE_{S_t,A^\A_t,S_{t+1} \sim\Pi}\left[\sum_{t=0} \gamma^t \ln \Pi(S_{t+1}|S_t,A^\A_t)\right]\Bigg)\nonumber\\
     &=(1-\gamma)\Bigg(\bbE_{S_t,A^\A_t,S_{t+1} \sim\Pi}\left[\sum_{t=0} \gamma^t Q^{\E}(S_{t+1}|S_t,A^\A_t)\right] - \gamma \left(\bbE_{S_{t+1}, S_t,A^\A_t\sim\Pi}\left[\sum_{t=1} \gamma^{t-1}Q^{\E}(S_{t+1}|S_t,A^\A_t) \right]\right)\nonumber\\
    &-  \gamma \left(\bbE_{S_{t+1}, S_t,A^\A_t\sim\Pi}\left[\sum_{t=1} \gamma^{t-1}\ln \Pi(S_{t+1}|S_t,A^\A_t) \right]\right) + \bbE_{S_t,A^\A_t,S_{t+1} \sim\Pi}\left[\sum_{t=0} \gamma^t \ln \Pi(S_{t+1}|S_t,A^\A_t)\right]\Bigg) \nonumber\\
    &=(1-\gamma) \bbE_{S_{1},S_0,A^\A_0}[Q^{\E}(S_{1}|S_0,A^\A_0) - \ln \Pi(S_{1}|S_0,A^\A_0)]\nonumber\\
    &=(1-\gamma) \bbE_{S_0} V^{\E}_\Pi(S_0).
\end{align}
Thus,
 \begin{align}
        \Theta(\Pi,Q^\E)  =  \bbE_{(S',A^\A,S)\sim \rho^\E_E}\!\! \left[{Q}^\E(S'|S,A^\A) \!-\! \gamma  {V}^\E_\Pi(S')\right] + (1-\gamma) \bbE_{S^0\sim P^0}{V}^\E_\Pi(S^0)
    \end{align}
    as desired.
\end{proof}

\subsection{Proof of Proposition \ref{th:th1}}
\textbf{Proposition \ref{th:th1}:}\textit{
   The problem   $\max_{Q^\E} \min_{\Pi^\E} J(\Pi^\E,Q^\E) $ is equivalent to the maximization 
$
{\max_{Q^\E}}J(\Pi^Q,Q^\E) 
$
where 
$$\Pi^Q(S'|S,A^\A) = \frac{\exp(Q^\E(S'|S,A^\A))}{ \sum_{S^{''}}\exp(Q^\E(S^{''}|S,A^\A))}
$$ 
Moreover, the function $J(\Pi^Q,Q^\E)$ is concave in $Q^\E$.} 
\begin{proof}
 To simplify the proof, we first consider the following simpler optimization problem. Let $p_1,p_2,...,p_N \in [0,1]$   and $x_1,...,x_N$ are N real numbers. Consider the maximization problem
\begin{align}
     \max_{p\in[0,1]^N} & \quad \sum_{i=1}^N x_ip_i - p_i\ln p_i\label{prob-ex}\tag{\sf P1} \\
    \mbox{subject to} &\quad \sum_{i} p_i =1 \nonumber
\end{align}
Using the KKT condition, if $p^*$ is an optimal solution to the above convex problem, $p^*$ needs to satisfy the following conditions: there exists $\lambda$ such that 
\[
\begin{cases}
  x_i - \ln p^*_i - 1 - \lambda = 0,~\forall i \\
 \sum_{i} p^*_i = 1
\end{cases}
\]
which implies 
\[
p^*_i = \exp(x_i-1-\lambda)
\]
Combine this with the condition $\sum_i p^*_i = 1$ we should have $\exp(-1-\lambda) = \sum_i \exp(x_i)$ and $p^*_i = \frac{\exp(x_i)}{\sum_{j}\exp(x_{j})}$ (*). In addition, when $p=p^*$, the objective function of \eqref{prob-ex} can be written as
\begin{align*}
f(x) &= \sum_{i}x_ip^*_i - p^*_i\ln(p^*_i)\\
&= \sum_{i}p_i\left(x_i - \ln(p^*_i)\right)\\
&= \sum_{i}p_i\left(x_i - x_i + \ln\left(\sum_i \exp(x_i)\right)\right)\\
& =\ln\left(\sum_i \exp(x_i)\right)
\end{align*}
We then see that $f(x)$ has a log-sum-exp form, thus it is convex in $x$  (**).

We now return to the minimax problem $\max_{Q^\E} \min_{\Pi} J(\Pi,Q^\E)$. For any fixed $Q^\E$, it can be seen that  the problem $\min_{\Pi} J(\Pi,Q^\E)$
is equivalent to
\[
\max_{\Pi} \Big\{ \gamma \bbE_{S}[V^\E_\Pi(S)]  + (1-\gamma) \bbE_{S^0\sim P^0}[V^\E_\Pi(S^0)]\Big\}
\]
We first consider the problem $\max_{\Pi} {V^\E (S)}$, recalling that $V^\E (S)$ can be written as
\begin{align*}
V^\E_\Pi(S) = &\sum_{A^\A} \Pi^\A(A^\A|S) \sum_{S'}\Pi(S'|S,A^\A)\times\left[Q^\E(S'|S,A^\A) - \log(\Pi(S'|S,A^\A))\right]
\end{align*}
Then it can be seen that each term $\sum_{S'}\Pi(S'|S,A^\A)
\times\left[Q^\E(S'|S,A^\A) - \log(\Pi(S'|S,A^\A))\right]$ has the form of \eqref{prob-ex}, thus from (**) we see that $V^\E_\Pi(S)$ is maximized at 
$$\Pi^Q(S'|S,A^\A) = \frac{\exp(Q^\E(S'|S,A^\A))}{ \sum_{S^{''}}\exp(Q^\E(s{''}_i|S,A^\A))}
$$ 

Moreover, according to the result (**) proved above, when $\Pi = \Pi^Q$, $V^\E_\Pi (S)$ is convex in $Q^\E$. Consequently,  the loss function of the IQ-learn  
\begin{align}
J(\Pi,Q^\E) \!=\!  \bbE_{(S',A^\A,S)\sim (\Pi_{\E},\Pi^\A)}\!\! \left[Q^\E(S'|S,A^\A) \!-\! \gamma  V^\E_\Pi(S')\right] -(1-\gamma) \bbE_{S^0\sim P^0}V^\E_\Pi(S^0) \nonumber
\end{align}
is concave in  $Q^\E$, which completes the proof. \textbf{Note that Corollary \ref{coro:1} is just a direct result from this proof.}
\end{proof}

\subsection{Proof of  Proposition \ref{prop:p1}}

\textbf{Proposition
\ref{prop:p1}:}\textit{
Given two allies' joint policies $\Pi^\A$ and $\widetilde{\Pi}^\A$ such that $\KL(\Pi^\A(\cdot|S) ||\widetilde{\Pi}^\A(\cdot|S))\leq \epsilon$ for any state $S\in \cS$, the following inequality holds:
{\small\begin{align}
&\left|\Phi(\Pi^\A|Q^\E) -  
\Phi(\widetilde{\Pi}^\A|Q^\E)\right| \leq  \left(\frac{\gamma+1+(1-\gamma)^3}{(1-\gamma)^2}\overline{Q} +\frac{1+(1-\gamma)^3}{(1-\gamma)^2} \ln |\cS| \right)\sqrt{2\ln 2 \epsilon  }\nonumber
\end{align}}
where $\overline{Q} \!=\! \!\max\limits_{(S'_{t},S_{t},A^\A_t)}\!\!Q^\E(S'_{t}|S_{t},A^\A_t)$ is an upper bound of $Q^\E$. }

\begin{proof}

We first write the loss function as 
\begin{align}
\Phi(\Pi^\A|Q^\E)  &\!=\!  \bbE_{(S',A^\A,S)\sim (\Pi_{\E},\Pi^\A)}\!\! \left[Q^\E(S'|S,A^\A) \!-\! \gamma  V^\E_Q(S'|\Pi^\A)\right] - (1-\gamma) \bbE_{S^0\sim P^0}V^\E_Q(S^0|\Pi^\A) \nonumber
\end{align}
where
\[
V^\E_Q(S|\Pi^\A) = \sum_{A^\A} \Pi^\A(A^\A|S) \ln\left(\sum_{S'}\exp\left(Q^\E(S'|S,A^\A)\right)\right).
\]
We first write the difference between two loss values as 
\begin{align}
    \Phi(\Pi^\A|Q^\E) -  \Phi(\widetilde{\Pi}^\A|Q^\E)  &\!=\! 
   \bbE_{(S',A^\A,S)\sim (\Pi_{\E},\Pi^\A)}\!\! \left[Q^\E(S'|S,A^\A)\right]  - \bbE_{(S',A^\A,S)\sim (\Pi_{\E},\widetilde{\Pi}^\A)}\!\! \left[Q^\E(S'|S,A^\A)\right] \nonumber
    \\
    &\!-\! 
   \left(\bbE_{(S',A^\A,S)\sim (\Pi_{\E},\Pi^\A)}\!\! \left[\gamma V^\E_Q(S'|\Pi^\A) \right]  - \bbE_{(S',A^\A,S)\sim (\Pi_{\E},\widetilde{\Pi}^\A)}\!\! \left[\gamma V^\E_Q(S'|\widetilde{\Pi}^\A) \right]\right) \nonumber
    \\
    &- \left((1-\gamma) \bbE_{S^0\sim P^0}(V^\E_Q(S^0|\Pi^\A) - V^\E_Q(S^0|\widetilde{\Pi}^\A))\right)\label{eq:lm1-eq11}
\end{align}

We first consider the first term of \eqref{eq:lm1-eq11}. Let us denote  the following function
\begin{align*}
    &\Gamma(S'|S,A^\A)= \bbE_{(S'_{t}|S_{t},A^\A_t) \sim ({\Pi}^\E,\widetilde{\Pi}^\A)}\Big[ \sum_{t=0}\gamma^t \left(Q^\E(S'_{t}|S_{t},A^\A_t)\right)\Big|(S'_{0}|S_{0},A^\A_0) = (S'|S,A^\A)\Big]
\end{align*}
and write the first term of \eqref{eq:lm1-eq11} as 
\begin{align}
&\bbE_{(S'_{t}|S_{t},A^\A_t) \sim ({\Pi}^\E,{\Pi}^\A)}\Big[ \sum_{t=0}\gamma^t \left(Q^\E(S'_{t}|S_{t},A^\A_t)\right)\Big] - \bbE_{(S'_{t}|S_{t},A^\A_t) \sim ({\Pi}^\E,\widetilde{\Pi}^\A)}\Big[ \sum_{t=0}\gamma^t \left(Q^\E(S'_{t}|S_{t},A^\A_t)\right)\Big]\nonumber\\
&=\bbE_{(S'_{t}|S_{t},A^\A_t) \sim ({\Pi}^\E,{\Pi}^\A)}\Big[ \sum_{t=0}\gamma^t \left(Q^\E(S'_{t}|S_{t},A^\A_t)\right)\Big] + \bbE_{(S'_{t}|S_{t},A^\A_t) \sim ({\Pi}^\E,{\Pi}^\A)}\Bigg[ \sum_{t=0}\gamma^t \Big(\gamma\Gamma(S'_{t+1}|S_{t+1},A^\A_{t+1}) \nonumber\\
&\qquad\qquad\qquad\qquad\qquad\qquad\qquad- \Gamma(S'_{t}|S_{t},A^\A_{t}) \Big)\Bigg] \nonumber\\
&=\bbE_{(S'_{t}|S_{t},A^\A_t) \sim ({\Pi}^\E,{\Pi}^\A)}\Big[ \sum_{t=0}\gamma^t \left(Q^\E(S'_{t}|S_{t},A^\A_t)+ \gamma\Gamma(S'_{t+1}|S_{t+1},A^\A_{t+1} - \Gamma(S'_{t}|S_{t},A^\A_{t})\right)\Big] \label{eq:lm1:eq2}
\end{align}

We also see that
\begin{align}
  \Gamma(S'_{t}|S_{t},A^\A_t)= Q^\E(S'_{t}|S_{t},A^\A_t) + \gamma\bbE_{(S'_{t+1},S_{t+1},A^\A_{t+1})\sim \Pi^\E,\widetilde{\Pi}^\A| S'_{t}} \Big[\Gamma(S'_{t+1}|S_{t+1},A^\A_{t+1})\Big]\nonumber
\end{align}
Thus, we further can bound the term \eqref{eq:lm1:eq2} as 
\begin{align*}
&\bbE_{(S'_{t+1},S_{t+1},A^\A_{t+1})\sim \Pi^\E,{\Pi}^\A| S'_{t}}\Big[ \left(Q^\E(S'_{t}|S_{t},A^\A_t)+ \gamma\Gamma(S'_{t+1}|S_{t+1},A^\A_{t+1} - \Gamma(S'_{t}|S_{t},A^\A_{t})\right)\Big] \\ &=\gamma\bbE_{(S'_{t+1},S_{t+1},A^\A_{t+1})\sim \Pi^\E,{\Pi}^\A| S'_{t}}\Big[ \left(\Gamma(S'_{t+1}|S_{t+1},A^\A_{t+1}\right)\Big] - \gamma \bbE_{(S'_{t+1},S_{t+1},A^\A_{t+1})\sim \Pi^\E,\widetilde{\Pi}^\A| S'}\Big[ \left(\Gamma(S'_{t+1}|S_{t+1},A^\A_{t+1}\right)\Big]  \\
   &\leq \gamma \cH ||\Pi^\A(\cdot|S_t)- \widetilde{\Pi}^\A(\cdot|S_t)||_\infty\nonumber \\
   &\leq \gamma \cH \max_{S} \left\{\sqrt{2\ln 2 \KL(\Pi^\A(\cdot|S) ||\widetilde{\Pi}^\A(\cdot|S) ) }\right\}
\end{align*}
where:
\begin{align*}
    &\cH \!= \!\max_{(S',S,A^\A)}\{ \Gamma(S'|S,A^\A)\} \\
    &= \!\max_{(S',S,A^\A)} \!\Bigg\{ \!\bbE_{(S'_{t}|S_{t},A^\A_t) \sim ({\Pi}^\E,\widetilde{\Pi}^\A)}\Big[ \sum_{t=0}\gamma^t \left(Q^\E(S'_{t}|S_{t},A^\A_t)\right)\Big]\!\Bigg\}\\
    & \leq \frac{\overline{Q}}{1-\gamma}
\end{align*}
where $\overline{Q}$ is an upper bound of $Q^\E(\cdot|\cdot,\cdot)$. Therefore, we can bound \eqref{eq:lm1:eq2} as 
\begin{align}\nonumber
   &\bbE_{(S'_{t}|S_{t},A^\A_t) \sim ({\Pi}^\E,{\Pi}^\A)}\Big[ \sum_{t=0}\gamma^t \left(Q^\E(S'_{t}|S_{t},A^\A_t)+ \gamma\Gamma(S'_{t+1}|S_{t+1},A^\A_{t+1} - \Gamma(S'_{t}|S_{t},A^\A_{t})\right)\Big]\\
    &\leq\! \frac{\gamma\overline{Q}}{(1-\gamma)^2}\max_{S} \left\{\sqrt{2\ln 2 \KL(\Pi^\A(\cdot|S) ||\widetilde{\Pi}^\A(\cdot|S) ) }\right\}\label{eq:lm1:b1}
\end{align}
For the second term of \eqref{eq:lm1-eq11}, we first bound
\begin{align}
    \left|V^\E_Q(S|\Pi^\A)  - V^\E_Q(S|\widetilde{\Pi}^\A) \right|&= \left|\sum_{A^\A} \Pi^\A(A^\A|S) \ln\left(\sum_{S'}\exp\left(Q^\E(S'|S,A^\A)\right)\right) - \sum_{A^\A} \Pi^\A(A^\A|S) \ln\left(\sum_{S'}\exp\left(Q^\E(S'|S,A^\A)\right)\right)\right|\nonumber\\
    &\leq (\ln|\cS| + \overline{Q}) \max_{S}||\Pi^\A(\cdot|S)- \widetilde{\Pi}^\A(\cdot|S)||_\infty\nonumber \\
    &\leq (\ln|\cS| + \overline{Q}) \max_{S} \left\{\sqrt{2\ln 2 \KL(\Pi^\A(\cdot|S) ||\widetilde{\Pi}^\A(\cdot|S) ) }\right\}\stackrel{def}{=} \cT \label{eq:bound-V}
\end{align}
Consequently, we can write the second term of \eqref{eq:lm1-eq11} as
\begin{align}
  &\left|\bbE_{(S',A^\A,S)\sim (\Pi_{\E},\Pi^\A)}\!\! \left[\gamma V^\E_Q(S'|\Pi^\A) \right]  - \bbE_{(S',A^\A,S)\sim (\Pi_{\E},\widetilde{\Pi}^\A)}\!\! \left[\gamma V^\E_Q(S'|\widetilde{\Pi}^\A) \right]\right|\nonumber\\
  &\leq \left|\bbE_{(S',A^\A,S)\sim (\Pi_{\E},\Pi^\A)}\!\! \left[\gamma V^\E_Q(S'|\Pi^\A) \right]  - \bbE_{(S',A^\A,S)\sim (\Pi_{\E},\widetilde{\Pi}^\A)}\!\! \left[\gamma V^\E_Q(S'|{\Pi}^\A) \right]\right| + \frac{\cT}{1-\gamma}\label{proof:eq12}
\end{align}
In a similar way, we write the first term of \eqref{proof:eq12} as 
\begin{align}\nonumber
   &  \left|\bbE_{(S',A^\A,S)\sim (\Pi_{\E},\Pi^\A)}\!\! \left[\gamma V^\E_Q(S'|\Pi^\A) \right]  - \bbE_{(S',A^\A,S)\sim (\Pi_{\E},\widetilde{\Pi}^\A)}\!\! \left[\gamma V^\E_Q(S'|{\Pi}^\A) \right]\right|\\\nonumber
   & =\left|\bbE_{(S_{t})\sim (\Pi_{\E},\Pi^\A)}\!\! \left[\sum_{t=0}\gamma^t V^\E_Q(S_{t}|\Pi^\A) \right]  - \bbE_{(S_{t})\sim (\Pi_{\E},\widetilde{\Pi}^\A)}\!\! \left[\sum_{t=0}\gamma^t V^\E_Q(S_{t}|{\Pi}^\A) \right]\right|\\\nonumber \\
   &=\bbE_{(S_{t} \sim {\Pi}^\E,\Pi^\A)}\!\Bigg[\! \sum_{t=0}\!\gamma^t\!\! \left(V^\E_Q(S_{t}|\Pi^\A) \!+\! \gamma U(S_{t+1})\!- \!U(S_{t})\!\right)\!\Bigg]\label{eq:lm1-eq4}
\end{align}
where $U(S) \!=\!  \bbE_{(S_{t} \sim {\Pi}^\E,\widetilde{\Pi}^\A)}\!\left[ \sum\limits_{t=0}\!\gamma^t\! \left(V^\E_Q(S_{t}|\Pi^\A)\right)\Big|~ s_{i0}\!=\! S\right]$. It then can be seen that
\begin{align}
    U(S_{t}) = V^\E_Q(S_{t}|\Pi^\A) + \gamma\bbE_{(S_{t+1})\sim \Pi^\E,\widetilde{\Pi}^A| S_{t}} [U(S_{t+1})]
\end{align}
Thus, given any $S_{t}$, we have
\begin{align}
   &\bbE_{S_{t+1}\sim (\Pi^\E,\Pi^\A)|S_{t}}\left[V^\E_Q(S_{t}|\Pi^\A) + \gamma U(S_{t+1})- U(S_{t})\right]\nonumber \\
   &=\gamma \bbE_{S_{t+1}\sim (\Pi^\E,\Pi^\A)|S_{t}}\left[ U(S_{t+1})\right] - \gamma \bbE_{S_{t+1}\sim (\Pi^\E,\widetilde{\Pi}^\A)|S_{t}}\left[ U(S_{t+1})\right]\nonumber \\
      &\leq \gamma \cK ||\Pi^\A(\cdot|S_t)- \widetilde{\Pi}^\A(\cdot|S_t)||_\infty\nonumber \\
   &\leq \gamma \cK  \max_{S} \left\{\sqrt{2\ln 2 \KL(\Pi^\A(\cdot|S) ||\widetilde{\Pi}^\A(\cdot|S) ) }\right\}
\end{align}
where $\cK = \max_{S} \{U(s_{i})\}$. We further see that
\begin{align*}
    &\max_{S} \{U(S)\} \leq \frac{1}{1-\gamma} \max_{S}V^\E_Q(S|\Pi^\A)\\
    &= \frac{1}{1-\gamma}\max_{S}\left\{\sum_{A^\A} \Pi^\A(A^\A|S) \ln\left(\sum_{S'}\exp\left(Q^\E(S'|S,A^\A)\right)\right).\right\} \\
    & \leq \frac{1}{1-\gamma}\max_{S,A^\A}\left\{\ln\left(\sum_{S'} \exp(Q^\E(S'|S,A^\A))\right)\right\}\\
    &\leq \frac{1}{1-\gamma}(\ln |\cS|+\overline{Q})
\end{align*}
So, \eqref{eq:lm1-eq4} can be bounded as 
\begin{align}
&\bbE_{(S_{t} \sim {\Pi}^\E,\Pi^\A)}\!\Bigg[\! \sum_{t=0}\!\gamma^t\!\! \left(V^\E_Q(S_{t}|\Pi^\A) \!+\! \gamma U(S_{t+1})\!- \!U(S_{t})\!\right)\!\Bigg] \nonumber\\
&\leq  \!\frac{\gamma(\ln |\cS|+\overline{Q})}{(1-\gamma)^2}\max_{S} \!\left\{\sqrt{2\ln 2 \KL(\Pi^\A(\cdot|S) ||\widetilde{\Pi}^\A(\cdot|S) ) }\right\}\label{eq:lm1:b2}
\end{align}
Combine \eqref{proof:eq12} and \eqref{eq:lm1:b2}, we bound the second term of \eqref{eq:lm1-eq11} as
\begin{align}
 &\left|\bbE_{(S',A^\A,S)\sim (\Pi_{\E},\Pi^\A)}\!\! \left[\gamma V^\E_Q(S'|\Pi^\A) \right]  - \bbE_{(S',A^\A,S)\sim (\Pi_{\E},\widetilde{\Pi}^\A)}\!\! \left[\gamma V^\E_Q(S'|\widetilde{\Pi}^\A) \right]\right|\nonumber\\
 &\leq \left(\frac{\gamma}{(1-\gamma)^2} + \frac{1}{1-\gamma}\right)(\ln |\cS|+\overline{Q}) \max_{S} \!\left\{\sqrt{2\ln 2 \KL(\Pi^\A(\cdot|S) ||\widetilde{\Pi}^\A(\cdot|S) ) }\right\} \nonumber \\
 &=\left(\frac{1}{(1-\gamma)^2} \right)(\ln |\cS|+\overline{Q}) \max_{S} \!\left\{\sqrt{2\ln 2 \KL(\Pi^\A(\cdot|S) ||\widetilde{\Pi}^\A(\cdot|S) ) }\right\} \label{term2}
\end{align}
The last term of \eqref{eq:lm1-eq11} can be bounded simply using \eqref{eq:bound-V}
as
\begin{align}
    \left|(1-\gamma) \bbE_{S^0\sim P^0}(V^\E_Q(S^0|\Pi^\A) - V^\E_Q(S^0|\widetilde{\Pi}^\A))\right|\leq (1-\gamma) \ln(|\cS| + \overline{Q}) \max_{S} \left\{\sqrt{2\ln 2 \KL(\Pi^\A(\cdot|S) ||\widetilde{\Pi}^\A(\cdot|S) ) }\right\}\label{term3}
\end{align}
Combine \eqref{eq:lm1:b1}, \eqref{term2} and \eqref{term3} we get 
\begin{align*}
    &\left|  \Phi(\Pi^\A|Q^\E) -  \Phi(\widetilde{\Pi}^\A|Q^\E)\right| \\
    &\leq \left(\frac{\gamma\overline{Q}}{(1-\gamma)^2} + \frac{\ln |\cS|+\overline{Q}}{(1-\gamma)^2} +  (1-\gamma) (\ln|\cS| + \overline{Q})\right)\max_{S} \left\{\sqrt{2\ln 2 \KL(\Pi^\A(\cdot|S) ||\widetilde{\Pi}^\A(\cdot|S) ) }\right\}\nonumber\\
    &=\left(\frac{\gamma+1+(1-\gamma)^3}{(1-\gamma)^2}\overline{Q} +\frac{1+(1-\gamma)^3}{(1-\gamma)^2} \ln |\cS| \right)\max_{S} \left\{\sqrt{2\ln 2 \KL(\Pi^\A(\cdot|S) ||\widetilde{\Pi}^\A(\cdot|S) ) }\right\}
\end{align*}
which completes the proof.
    
\end{proof}

\subsection{Proof of Proposition \ref{prop:bound-continous}}
\textbf{Proposition \ref{prop:bound-continous}:} 
\textit{Given two allies' joint policies $\Pi^\A$ and $\widetilde{\Pi}^\A$ such that for every state $S\in \cS$, $\KL(\Pi^\A(\cdot|S) ||\widetilde{\Pi}^\A(\cdot|S))\leq \epsilon$, the following inequality holds:
{\small\begin{align}
&\left|\Phi(\Pi^\A|Q^\E,\Pi) -  
\Phi(\widetilde{\Pi}^\A|Q^\E ,\Pi)\right| \nonumber\\
&\qquad\leq  \left(\frac{\gamma+1+(1-\gamma)^3}{(1-\gamma)^2}\overline{Q}  - \frac{1+(1-\gamma)^3}{(1-\gamma)^2} H \right)\sqrt{2\ln 2 \epsilon  }\nonumber
\end{align}}
where $\overline{Q}=\sup\limits_{(S'_{t},S_{t},A^\A_t)}\!\!Q^\E(S'_{t}|S_{t},A^\A_t)$ is an upper bound of $Q^\E(\cdot|\cdot,\cdot)$ and $H = \inf\limits_{(S,A^\A)}\sum_{S}\Pi(S'|S,A^\A)\ln \Pi(S'|S,A^\A)$, is a lower bound of the entropy of the actor policy $\Pi(\cdot|\cdot,\cdot)$.   } 

\begin{proof}
   We can follow the same arguments as in the proof of Proposition \ref{prop:p1} above, with the only difference being when we bound $V^\E_Q(S)$. Here, $V^\E_Q(S)$ is replaced by $V^\E_\Pi(S)$ and can be bounded as
   \begin{align}
      V^\E_\Pi(S)&= \sum_{A^\A} \Pi^\A(A^\A|S) \sum_{S'}\Pi(S'|S,A^\A)\left[Q^\E(S'|S,A^\A) - \log(\Pi(S'|S,A^\A))\right] \nonumber \\
      &\leq \max_{Q^\E, \Pi} \{Q^\E(S'|S,A^\A)\} - \min_{\Pi}\{\Pi(S'|S,A^\A)\log(\Pi(S'|S,A^\A))\} \nonumber \\
      &\leq \overline{Q} - {H}      
   \end{align}
   where ${H} = \inf_{S,A^\A} \{H(S,A^\A)\}$, and 
 $H(S,A^\A)$ is the entropy of $\Pi(\cdot|S,A^
\A)$, i.e., $\sum_{S,A^\A}\Pi(S'|S,A^
\A) \ln \Pi(S'|S,A^
\A)$. The difference between $V^\E_\Pi(S|\Pi^\A)$ and $V^\E_\Pi(S|\widetilde{\Pi}^\A)$ \footnote{Here we write $V^\E_\Pi(S)$ as a function of $\Pi^\A$} can be bounded as
\begin{align}
    |V^\E_\Pi(S|\Pi^\A) - V^\E_\Pi(S|\widetilde{\Pi}^\A)|&\leq  \max_{Q^\E, \Pi} \{\sum_{S'}\Pi(S'|S,A^
\A)(Q^\E(S'|S,A^\A) - \log(\Pi(S'|S,A^\A)))\} \times \max_S||\Pi^\A(\cdot|S)- \widetilde{\Pi}^\A(\cdot|S)||_\infty\nonumber\\
    &\leq  (-H + \overline{Q}) \max_{S} \left\{\sqrt{2\ln 2 \KL(\Pi^\A(\cdot|S) ||\widetilde{\Pi}^\A(\cdot|S) ) }\right\}
\end{align}
 Therefore, the overall bound can be established by replacing the term $\ln |\cS|$ in the discrete case by $-H$. We then obtain
\begin{align}
\left|\Phi(\Pi^\A|Q^\E,\Pi) -  
\Phi(\widetilde{\Pi}^\A|Q^\E ,\Pi)\right| \leq  \left(\frac{\gamma+1+(1-\gamma)^3}{(1-\gamma)^2}\overline{Q} -\frac{1+(1-\gamma)^3}{(1-\gamma)^2} H \right)\sqrt{2\ln 2 \epsilon  }\nonumber
\end{align}
\end{proof}

\subsection{Proof of Corollary \ref{coro.1}}
\textbf{Corollary \ref{coro.1}:}\textit{
Given two allies' joint policies $\Pi^\A$ and $\widetilde{\Pi}^\A$ with $\forall S\!\in\! \cS$, $\KL(\Pi^\A(\cdot|S) ||\widetilde{\Pi}^\A(\cdot|S)\!\leq\! \epsilon$, the following holds:
{\small\begin{align*}
&\Big|\max\nolimits_{Q^\E}\!\!\big\{\Phi(\Pi^\A|Q^\E)\big\} -  
\max\nolimits_{Q^\E}\!\!\big\{\Phi(\widetilde{\Pi}^\A|Q^\E)\big\}\Big| \leq \cO(\sqrt{\epsilon}) \;\text{ (discrete)}\\   
&\Big|\max\nolimits_{Q^\E}\min\nolimits_\Pi\big\{\Phi(\Pi^\A|Q^\E,\Pi)\big\}-  
\max\nolimits_{Q^\E} \min\nolimits_{\Pi}\big\{\Phi(\widetilde{\Pi}^\A|Q^\E,\Pi)\big\}\Big| \leq \cO(\sqrt{\epsilon})\qquad\text{ (continuous)}
\end{align*}}}

\begin{proof}
To simplify the notation, we first prove the following  result: 
\begin{lemma}\label{lm-coro}
Given $\epsilon>0$, and two functions $f(x)$, $g(x)$ such that $|f(x)-g(x)|\leq \epsilon$ for any $x\in \cX$ ($\cX$ is the feasible set of $x$). The following hold true
\begin{align*}
    |\max_x f(x) - \max_x g(x)| &\leq \epsilon \\
    |\min_x f(x) - \min_x g(x)| &\leq \epsilon
\end{align*}  
\end{lemma}
To prove the above lemma, let $x^f,x^g$ be optimal solutions to $\max_x f(x)$ and $\min_x g(x)$, respectively.  We consider 2 cases
\begin{itemize}
    \item [(i)] If $\max_x f(x) \geq \min_x g(x)$, then we have
    \begin{align*}
        |\max_x f(x) - \min_x g(x)| = f(x^f) - g(x^g) \leq f(x^f) - g(x^f) \stackrel{(a)}{\leq} \epsilon
    \end{align*}
    where $(a)$ is because $|f(x)-g(x)|\leq \epsilon$ for any $x\in\cX$
    \item [(ii)] If $\max_x f(x) \leq \min_x g(x)$, then we have
    \begin{align*}
        |\max_x f(x) - \min_x g(x)| = g(x^g) - f(x^f)  \leq g(x^g) - f(x^g) {\leq} \epsilon
    \end{align*}
\end{itemize}
Thus $|\max_x f(x) - \max_x g(x)| \leq \epsilon $. The inequality  $|\min_x f(x) - \min_x g(x)| \leq \epsilon$ can be validated in the same way. 

We now get back to the main proof. Since $\left|\Phi(\Pi^\A|Q^\E) -  
\Phi(\widetilde{\Pi}^\A|Q^\E)\right|\leq \cO(\sqrt{\epsilon}) $ (Proposition \ref{prop:p1}), Lemma \ref{lm-coro} above implies that
\[
\Big|\max\nolimits_{Q^\E}\!\!\big\{\Phi(\Pi^\A|Q^\E)\big\} -  
\max\nolimits_{Q^\E}\!\!\big\{\Phi(\widetilde{\Pi}^\A|Q^\E)\big\}\Big| \leq \cO(\sqrt{\epsilon})
\]
Moreover, from Proposition \ref{prop:bound-continous}, applying Lemma \ref{lm-coro} twice, we have the following
\[
\Big|\min\nolimits_\Pi\big\{\Phi(\Pi^\A|Q^\E,\Pi)\big\}-  
\min\nolimits_{\Pi}\big\{\Phi(\widetilde{\Pi}^\A|Q^\E,\Pi)\big\}\Big| \leq \cO(\sqrt{\epsilon})
\]
and
\[
\Big|\max\nolimits_{Q^\E}\min\nolimits_\Pi\big\{\Phi(\Pi^\A|Q^\E,\Pi)\big\}-  
\max\nolimits_{Q^\E} \min\nolimits_{\Pi}\big\{\Phi(\widetilde{\Pi}^\A|Q^\E,\Pi)\big\}\Big| \leq \cO(\sqrt{\epsilon})
\]
which completes the proof.
\end{proof}

\section{Other Experimental Settings and Results}

For a fair comparison between our proposed algorithm and existing methods, we use the same model architecture and hyperparameters as shown in Tables \ref{hyper} and \ref{model} respectively. We also use Value Normalization method and Recurrent Neural Network mechanism proposed by MAPPO \cite{yu2022surprising}, which used as an improvement method to speed up the training process.
All sub-tasks (SMACv2, GRF, Miner) are trained concurrently in a GPU-accelerated HPC (High Performance Computing). Therefore, running times reported might not be accurate. In average, each sub-task requires 4-7 days of training depending on its difficulty. As a result, due to the limitation of our computing resources, we do not test our method's performance with other settings such as disabling Value Normalization, using MLP instead of Recurrent Neural Network, tuning learning rate, tuning clip range, etc.

\begin{figure}[htb]
\centering
\showlegend[0.18]{24.7}{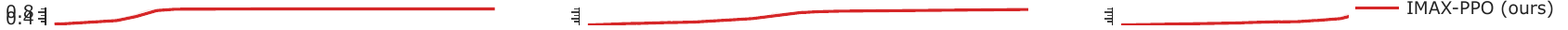} \\

\showinstance[0.16]{0}{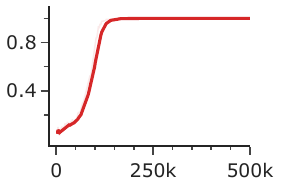}{3_vs_1}
\showinstance[0.14]{18}{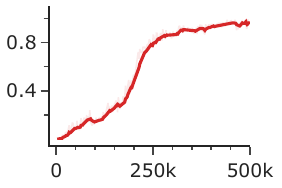}{counter_easy}
\showinstance[0.14]{18}{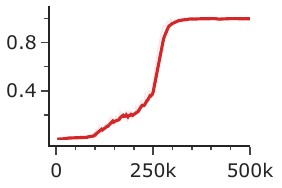}{counter_hard}

\caption{Win-rate curves on  GRF environment.}
\label{fig:grf}
\end{figure}

\begin{figure}[htb]
\centering
\showlegend[0.35]{20.2}{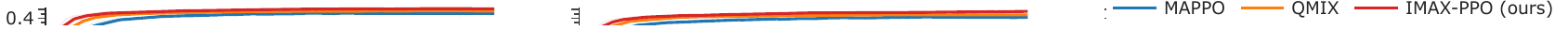} \\

\showinstance[0.16]{0}{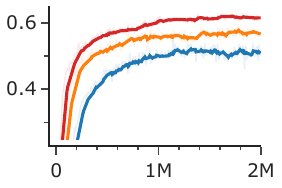}{easy_2_vs_2}
\showinstance[0.14]{18}{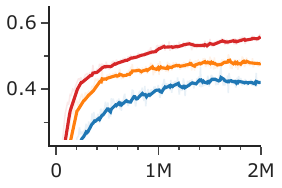}{medium_2_vs_2}
\showinstance[0.14]{18}{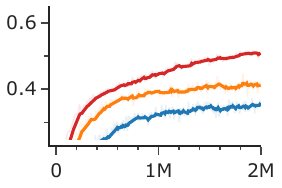}{hard_2_vs_2}

\caption{Win-rate curves on Gold Miner environment.}
\label{miner-cur}
\end{figure}

Regarding GRF tasks, there are 11 academy scenarios (depicted at \url{https://github.com/google-research/football/blob/master/gfootball/doc/scenarios.md}). Although testing in all these sub-tasks is interesting, but we only evaluate on top three importance scenarios: \textit{\detokenize{academy_3_vs_1_with_keeper}}, \textit{\detokenize{academy_counterattack_easy}}, and \textit{\detokenize{academy_counterattack_hard}}. All are the hardest sub-tasks with the highest number of agents except the \textit{almost} full football scenarios \textit{\detokenize{academy_single_goal_versus_lazy}}. We report the  win-rate curves of our IMAX for the three tasks in Figure \ref{fig:grf}. We do not show the performance curves of the other baselines methods (QMIX and MAPPO) as they are not available in their papers, noting that the final win-rates of all the methods considered are already reported in the main paper. 
Finally, Figure \ref{miner-cur} shows the win-rate curves on the Gold Miner tasks.

\twocolumn

\begin{table}[htp]
\centering
\begin{tabular}{c|ccc}
                    & SMACv2       & GRF       & Miner   \\ \hline
Runner              & \multicolumn{3}{c}{Parallel}       \\
Workers             & \multicolumn{2}{c}{8}    & 32      \\
Total steps         & 10e6         & 0.5e6     & 2e6     \\
Mini steps          & \multicolumn{3}{c}{1024}           \\
Evaluation          & \multicolumn{3}{c}{32}             \\
Mini batch          & \multicolumn{3}{c}{1}              \\
Device              & \multicolumn{3}{c}{Cuda}           \\
Mini epochs         & \multicolumn{3}{c}{5}              \\
Learning rate       & \multicolumn{2}{c}{5e-4} & 2e-4    \\
Epsilon                 & \multicolumn{3}{c}{1e-5}           \\
Weight decay        & \multicolumn{3}{c}{0.0}            \\
Clip range          & \multicolumn{3}{c}{0.2}            \\
Entropy coefficient & 0.01         & 0.0       & 0.01    \\
Value coefficient   & \multicolumn{3}{c}{0.5}            \\
Weight gain         & \multicolumn{3}{c}{0.01}           \\
Weight initializer  & \multicolumn{3}{c}{Orthogonal}     \\
Max gradient norm   & \multicolumn{3}{c}{10.0}           \\
Gamma (GAE)         & \multicolumn{3}{c}{0.99}           \\
Lambda (GAE)        & \multicolumn{3}{c}{0.95}    \\ \hline     
\end{tabular}
\caption{Hyperparameters}
\label{hyper}
\end{table}

\begin{table}[htp]
\centering
\begin{tabular}{|c|c|c|}
\hline
\multirow{8}{*}{Actor}  & \multirow{4}{*}{Base}        & LayerNorm \\
&& Linear    \\
&& ReLU      \\
&& LayerNorm \\ \cline{2-3}
& \multirow{2}{*}{RNN}         & GRU       \\
&& LayerNorm \\ \cline{2-3}
& \multirow{2}{*}{Categorical} & Linear    \\
&& Masking   \\ \hline \hline
\multirow{7}{*}{Critic} & \multirow{4}{*}{Base}        & LayerNorm \\
&& Linear    \\
&& ReLU      \\
&& LayerNorm \\ \cline{2-3}
& \multirow{2}{*}{RNN}         & GRU       \\
&& LayerNorm \\ \cline{2-3}
& Value& Linear   \\ \hline
\end{tabular}
\caption{PPO model architecture}
\label{model}
\end{table}

\end{document}